\documentclass[final,5p,times,twocolumn]{elsarticle}
\usepackage[utf8]{inputenc}
\usepackage{array}
\usepackage{lscape} 
\usepackage{acronym}
\usepackage{amsmath}
\usepackage{graphicx}
\usepackage{hyperref}
\usepackage{paralist}
\usepackage{rotating}
\usepackage{longtable}
\usepackage{url}
\usepackage{tikz}
\usepackage{caption}
\usepackage{subcaption}

\acrodef{ABED}[ABED]{Algorithm-Based Error Detection}
\acrodef{ABFT}[ABFT]{Algorithm-Based Fault Tolerance}
\acrodef{AI}[AI]{Artificial Intelligence}
\acrodef{ANN}[ANN]{Artificial Neural Network}
\acrodef{BNN}[BNN]{Binarized Neural Network}
\acrodef{CNN}[CNN]{Convolutional Neural Network}
\acrodef{DL}[DL]{Deep Learning}
\acrodef{DNN}[DNN]{Deep Neural Network}
\acrodef{DUE}[DUE]{Detected Unrecoverable Error}
\acrodef{DWC}[DWC]{Duplication with Comparison}
\acrodef{ECC}[ECC]{Error Correcting Code}
\acrodef{MCU}[MCU]{Multiple Cell Upsets}
\acrodef{NMR}[NMR]{N-Modular Redundancy}
\acrodef{SA}[SA]{Stuck-at}
\acrodef{SDC}[SDC]{Silent Data Corruption}
\acrodef{SEU}[SEU]{Single Event Upset}
\acrodef{SET}[SET]{Single Event Transient}
\acrodef{TMR}[TMR]{Triple Modular Redundancy}
\acrodef{TF}[TF]{TensorFlow}
\acrodef{ML}[ML]{Machine Learning}
\acrodef{GPU}[GPU]{Graphic Processing Unit}
\acrodef{FPGA}[FPGA]{Field Programmable Gate Array}
\acrodef{RTL}[RTL]{Register-Transfer Level}
\acrodef{ISA}[ISA]{Instruction Set Architecture}
\acrodef{MBU}[MBU]{Multiple Bit Upset}
\acrodef{LSB}[LSB]{Least Significant Bit}
\acrodef{TPU}[TPU]{Tensor Processing Unit}
\acrodef{MAC}[MAC]{Multiply and Accumulate}
\acrodef{IR}[IR]{Intermediate Representation}
\acrodef{HPC}[HPC]{High Performance Computing}
\acrodef{PVF}[PVF]{Program Vulnerability Factor}
\acrodef{KVF}[KVF]{Kernel Vulnerability Factor}
\acrodef{FIT}[FIT]{Failures in Time }
\acrodef{MTTF}[MTTF]{Mean Time To Failure}
\acrodef{AVF}[AVF]{Architecture Vulnerability Factor}
\acrodef{PE}[PE]{Processing Element}

\newcommand{\nuovo}[1]{\textcolor{black}{#1}} 

\newcommand{\cir}[1]{\tikz[baseline]{%
    \node[anchor=base, draw, circle, inner sep=0, minimum width=1.2em]{#1};}}

\def\numsearchoutput{2165}
\def\numpre{242}
\def\numread{220}
\def\numincluded{84}
\def\numintabs{76}

\journal{Computer Science Review}

\begin{document}

\begin{frontmatter}

\title{Resilience of Deep Learning applications: a systematic literature review of analysis and hardening techniques}


\author[1]{Cristiana Bolchini\corref{cor1}} 
\ead{cristiana.bolchini@polimi.it}
\author[1]{Luca Cassano}
\ead{luca.cassano@polimi.it}
\author[1]{Antonio Miele}
\ead{antonio.miele@polimi.it}

\date{May 9, 2024}

\affiliation[1]{organization={Politecnico di Milano, Dip. Elettronica, Informazione e Bioingegneria},
            addressline={P.zza L. da Vinci, 32}, 
            city={Milan},
            postcode={20133}, 
            state={},
            country={Italy}}

\cortext[cor1]{Corresponding author}

\begin{abstract}
\ac{ML} is currently being exploited in numerous applications being one of the most effective \ac{AI} technologies, used in diverse fields, such as vision, autonomous systems, and alike. The trend motivated a significant amount of contributions to the analysis and design of \ac{ML} applications against faults affecting the underlying hardware. The authors investigate the existing body of knowledge on \acl{DL} (among \ac{ML} techniques) resilience against hardware faults systematically through a thoughtful review in which the strengths and weaknesses of this literature stream are presented clearly and then future avenues of research are set out. The review is based on \numread{} scientific articles published between January 2019 and March 2024.
The authors adopt a classifying framework to interpret and highlight research similarities and peculiarities, based on  several parameters, starting from the  main scope of the work, the adopted fault and error models, to their reproducibility. This framework allows for a comparison of the different solutions and the identification of possible synergies. Furthermore, suggestions concerning the future direction of research are proposed in the form of open challenges to be addressed.
\end{abstract}





\end{frontmatter}

\acresetall

\section{Introduction}
The widespread adoption of \ac{ML} in safety/mission-critical systems motivated a great attention towards the resilience of such complex systems against the occurrence of faults in the underlying hardware. 
Among all \ac{ML} techniques, \ac{DL} is the one that the research community is mainly focusing its attention on, also in terms of reliability issues. In fact, \ac{DL} is widely used for vision and perception functionalities, which are particularly relevant for implementing human-assisting tasks (e.g., advanced driver-assistance systems) and represent the enabling technology for autonomous behaviors (e.g., unmanned aerial vehicles or rovers). 
\ac{DL} consists of a set of specific \ac{ANN} models where multiple layers of processing are used to extract progressively higher level features and information from raw data, such as images taken from cameras. 
In general, faults can occur in \begin{inparaenum} \item input data, \item software, and \item hardware, \end{inparaenum} possibly causing the \ac{DL} application to behave differently from what expected (e.g., \cite{WS+2022}). Faults on \textit{input data} may derive from defective/broken sensors and devices, noise, as well as from adversarial attacks. Faults in \textit{software} usually originate from bugs or aggressive implementations. Finally, faults in \textit{hardware} may be caused by radiation, voltage over-scaling, and aging or in-field permanent stuck-at failures. When addressing with hardware faults, the underlying assumption is that the \ac{DL} application has been designed and implemented to achieve the best performance (in terms of accuracy of the prediction tasks) with respect to requirements and constraints, and the input data is genuine.  
In this work we focus on \textit{hardware faults} and we investigate analysis and design methods and tools to evaluate and possibly improve the \textit{resilience} of \ac{DL} algorithms and applications against this source of failure. Sometimes the term \textit{robustness} is interchangeably used, still referring to the mentioned hardware faults context; we do not investigate \textit{resilience/robustness} with respect to its design and implementation, nor to adversarial attacks, belonging to the security scenario. 
Moreover, although the design and training processes have an impact on the performance of the final implementation resilience, such facets are here considered only when they are tailored to the possibility to mitigate hardware fault effects. 

On this topic the body of knowledge is quite rich, and a very detailed analysis has been presented in \cite{Mit2020}, where the author introduces a comprehensive and extensive synthesis of analysis and hardening methods against faults affecting hardware platforms running \ac{ANN} applications. The contribution details the various adopted fault models, the fault simulation/injection and emulation strategies presented in literature at that time, as well as the proposed solutions to make the \ac{AI} resilient against the analyzed faults/errors.  
A similar contribution is given by \cite{IW+2020}, where the authors analyze how faults in \ac{DNN} accelerators such as \acp{GPU} and \acp{FPGA}, affect the executed application. The analysis framework takes into account the different sources of faults and possible fault locations, and a few final considerations mention hardening solutions. The most recent contribution reporting part of the body of work on \ac{DL} resilience is \cite{RSL+2023}, analyzing some recent research and results focused on resilience assessment. The authors introduce the context and detail the fault analysis strategies and methods adopted when dealing with \ac{DL} applications, reporting some novel solutions. These papers serve not only as a reference to prominent research up to that time instant, but also provide a concise explanation of the various existing techniques. To complete the scenario overview, three recent contributions that briefly discuss the state of the art and focus also on possible research challenges and opportunities are the works presented in~\cite{ZL+2019, HS2020_1, WS+2022}, sometimes embracing also security-related considerations. 

\begin{figure}[t]
    \centering
    \includegraphics[width=\columnwidth]{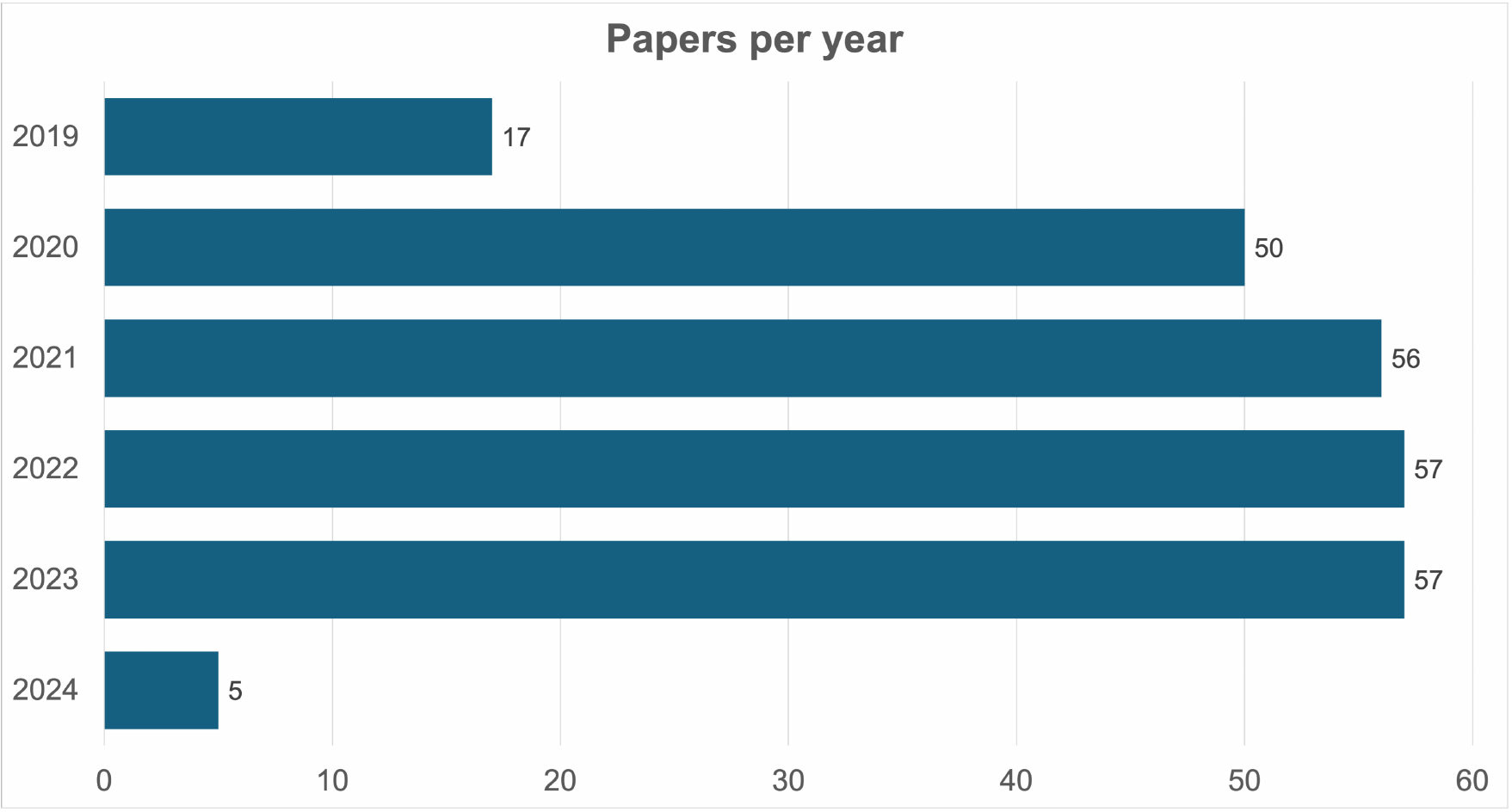}
    \caption{Number of contributions on the domain of interest per year, in the considered time frame.}
    \label{fig:timeline}
\end{figure}

As Figure~\ref{fig:timeline} shows, the community is very active and the contributions of the last four years introduce new relevant elements and insights, motivating, in our opinion, a new review, that also introduces different perspectives with respect to recent ones (i.e., \cite{RSL+2023}); a classification framework as well as other synthetic considerations on the observed trends. 

Given the breadth of the domain and the many different facets, we define a boundary based on \begin{inparaenum}
\item the time window of the publication, selecting only those included in the Jan. 2019 - \nuovo{Mar. 2024} window, with a couple of exceptions mainly referring to surveys (e.g., \cite{RSL+2023}) and tools still commonly adopted to perform resilience analysis (e.g., \cite{RG+2018}), to better frame the discussion;
\item the \textit{adopted fault model}, by including only contributions that cover transient and permanent faults, excluding those that only affect weights stored in memories that are typically protected by \ac{ECC} solutions;
\item the \textit{\ac{DL} algorithm}, by excluding works that strictly depend on the specific \ac{ANN} architecture (e.g., spiking, \nuovo{vision transformers}), so that the presented solutions can be broadly adopted;
\item the hardware platform running the application, by including CPUs and hardware accelerators, such as \acp{GPU} and \acp{FPGA}.
\end{inparaenum}

The rest of the paper is organized as follows. Section~\ref{sec:method} introduces the adopted search methodology aligned with the boundary of the domain previously mentioned, and the classification framework defined to analyze the available contributions. Section~\ref{sec:soa} reports the various contributions, characterized according to the defined analysis framework, briefly summarizing the most relevant aspects. Section~\ref{sec:insights} draws some considerations on the overall state of the art, highlighting open challenges and opportunities, while Section~\ref{sec:end} concludes the paper.

\section{Methodology}\label{sec:method}
Before presenting the proposed classification framework and the selected contributions, we here present the adopted search and selection process.

\subsection{Research design}
This study aims at conducting a systematic literature review to explore the current state of the art in the design and analysis of resilient \ac{DL} applications against hardware faults and to observe the present research trends in this context. The purpose is to get an up-to-date overview of the available solutions, also identifying the open challenges and possible opportunities in the field. To this end we performed a thorough search and designed an analysis framework to classify the numerous contributions. 

\subsection{Research method}
To gather the contributions within the area of interest, we started from Scopus and World of Science to collect papers that appeared in renowned venues (both journals and conferences), delimiting the time span between January 2019 and March 2024, and excluding all topic areas and keywords that would surely lead to not relevant publications. Tables~\ref{tab:searchstrings} and \ref{tab:searchdbstrings} report the desired search strings and the actual ones in the mentioned repositories. 

\begin{table*}[t]
\caption{The implemented search strings.}\label{tab:searchstrings}
\begin{center}
\begin{tabular}{m{2em} m{13cm} }
  \hline
  1 & ("soft error" OR "resilien*" OR "dependab*" OR "fault toleran*" OR "reliab*" OR robust OR "harden*")  \\
  2 & ("\acl{DL}" OR \acs{DL} OR "\acl{ML}" OR \acs{ML}) \\
  3 & ("\acl{CNN}" OR "\acl{CNN}" OR \acs{CNN} OR "\acl{DNN}" OR \acs{DNN}) \\
  4 & ("soft error" OR fault OR "\acl{SEU}" OR \acs{SEU}) \\
  \hline
\end{tabular}
\end{center}
\end{table*}

\begin{table*}[t]
\caption{The selected databases and formulated search strings.}\label{tab:searchdbstrings}
\begin{center}
\begin{tabular}{lm{13cm}}
  \hline
  \textbf{Database} & \textbf{Search string} \\
  \hline
  Scopus & TITLE-ABS-KEY ( ( "Resilien*" OR "Fault toleran*" OR "Robust*" OR "Dependab*" OR "Reliab*" ) AND ( "CNN" OR "DNN" OR ml OR "Convolutional Neural Network" OR "Deep Neural Network" ) AND ( "Soft error" OR seu OR fault ) ) AND PUBYEAR > 2018 AND ( EXCLUDE ( SUBJAREA,"PHYS" ) OR EXCLUDE ( SUBJAREA,"MATH" ) OR EXCLUDE ( SUBJAREA,"ENER" ) OR EXCLUDE ( SUBJAREA,"MATE" ) OR EXCLUDE ( SUBJAREA,"DECI" ) OR EXCLUDE ( SUBJAREA,"CHEM" ) OR EXCLUDE ( SUBJAREA,"EART" ) OR EXCLUDE ( SUBJAREA,"BIOC" ) OR EXCLUDE ( SUBJAREA,"CENG" ) OR EXCLUDE ( SUBJAREA,"ENVI" ) OR EXCLUDE ( SUBJAREA,"MULT" ) OR EXCLUDE ( SUBJAREA,"SOCI" ) OR EXCLUDE ( SUBJAREA,"NEUR" ) OR EXCLUDE ( SUBJAREA,"MEDI" ) OR EXCLUDE ( SUBJAREA,"BUSI" ) OR EXCLUDE ( SUBJAREA,"HEAL" ) OR EXCLUDE ( SUBJAREA,"AGRI" ) ) AND ( EXCLUDE ( LANGUAGE,"Chinese" ) OR EXCLUDE ( LANGUAGE,"French" ) ) AND ( EXCLUDE ( EXACTKEYWORD,"Diagnos" ))
   \\
  \hline
  WOS &  ((TI=("Resilien*" OR "Fault toleran*" OR "Robust*" OR "Dependab*" OR "Reliab*") OR AK=("Resilien*" OR "Fault toleran*" OR "Robust*" OR "Dependab*" OR "Reliab*")) AND (TI=("CNN" OR "DNN" OR ML OR "Convolutional Neural Network" OR "Deep Neural Network" OR ML OR "Machine Learning" OR DL OR "Deep Learning") OR AK=("CNN" OR "DNN" OR ML OR "Convolutional Neural Network" OR "Deep Neural Network" OR ML OR "Machine Learning" OR DL OR "Deep Learning")) AND (TI=("Soft error" OR SEU OR fault) OR AK=("Soft error" OR SEU OR fault))) \\
  \hline
\end{tabular}
\end{center}
\end{table*}

\begin{figure}[t]
    \centering
    \includegraphics[width=\columnwidth]{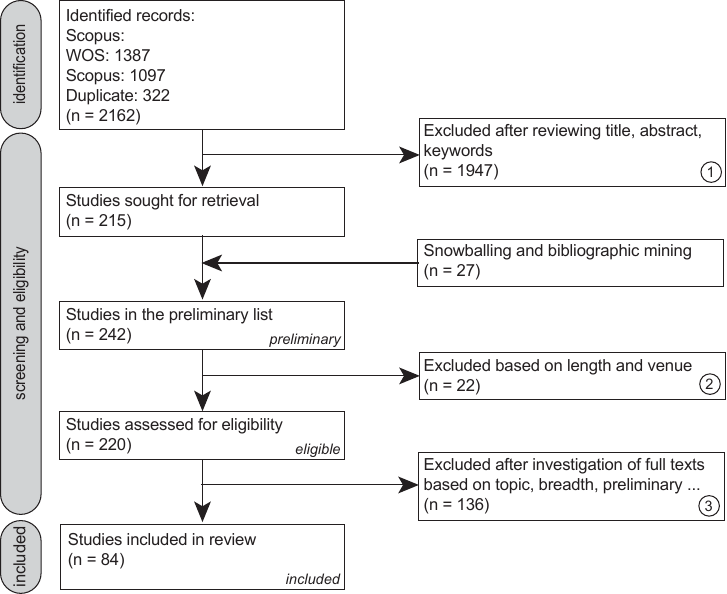}
    \caption{Flow diagram presenting the retrieval and screening process of the literature following the Preferred Reporting Items for Systematic Reviews and Meta‐Analyses (PRISMA) process.}
    \label{fig:prisma}
\end{figure}

The searches returned a very high number of contributions (\numsearchoutput{}) and we adopted the process reported in Figure~\ref{fig:prisma} to filter out clearly unrelated contributions and to include other ones through reference mining and snowballing also on other search engines. More precisely, we initially excluded contributions (filter $\cir{1}$) based on the title, the abstract and the keywords. Indeed many results referred to the use of \ac{ML}/\ac{DL} for resilience and diagnosis, sometimes applied to out-of-scope contexts (e.g., power/transmission lines or not \ac{ML}/\ac{DL} applications). Through snowballing and reference mining we added new contributions, leading to a batch of \numpre{} papers we read. Further filtering took place (filter $\cir{2}$) based on the strength of the contribution (length and/or venue) and the existence of a subsequent more mature/complete publication (\numread{} papers, dubbed \textit{eligible}). Finally, we selected a set of \numincluded{} papers considered as the review sources (filter $\cir{3}$) to have contributions presenting solutions of general validity, possibly excluding too specific scenarios or custom case study. 6 out of the \numincluded{} documents are surveys or position papers, 2 are tools not specific to resilience analysis/hardening, thus we actually analyze and classify \numintabs{} papers, presenting novel contributions on the topic of interest.
The characteristics of the search method as well as the outcomes are summarized in Table~\ref{tab:search}. 
The spreadsheet file with all the raw bibliographic data analyzed during this systematic literature review process can be downloaded from \url{https://github.com/D4De/dl\_resilience\_survey}.

\begin{table*}[t]
\caption{Search methodology details.}\label{tab:search}
\begin{center}
\begin{tabular}{ | m{4.2cm} | m{10cm} | }
  \hline
  Keywords: & soft error, resilience, dependable, fault tolerance, reliable, robust AND \\
  & \ac{DL}, \acsp{CNN}, \acp{DNN} \\
  \hline
  Repositories: &  IEEE, ACM, Elsevier, Springer \\ 
  \hline
  Search engines: & Google scholar, Semantic scholar, Scopus, lens.org, DBLP \\ 
  \hline
  Publication years: & January 2019 - March 2024 \\ 
  \hline
  Search outcome:  &  \numsearchoutput \\ 
  \hline
  Analyzed contributions:  &  \numread \\ 
  \hline
  Reported contributions: &  \numincluded \\  
  \hline
  Novel technical contributions: &  \numintabs \\  
  \hline
\end{tabular}
\end{center}
\end{table*}

\subsection{Classification framework}
\begin{figure*}[htp]
    \centering
    \includegraphics[width=\textwidth]{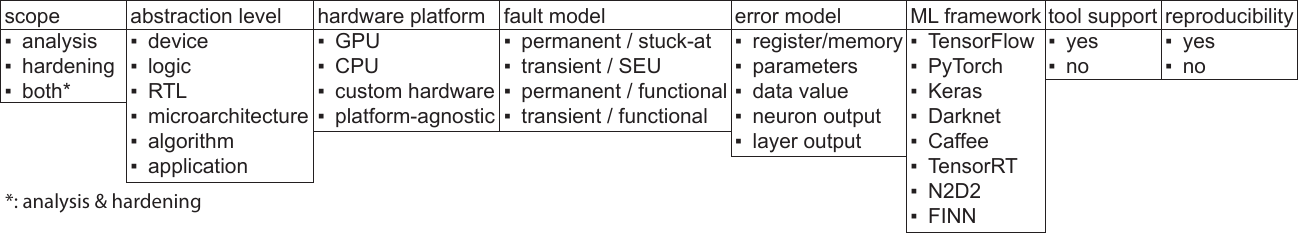}
    \caption{The primary axes of the adopted classification framework, with a few sample values.}
    \label{fig:framework}
\end{figure*}

We have defined an analysis framework to carry out a rigorous classification of the selected papers. Figure~\ref{fig:framework} sketches the primary axes of this analysis framework, being a set of relevant aspects for the considered topic, i.e., system's \textit{resilience}, and the referred application scenario, i.e., \textit{\ac{DL} applications}. A brief description of all the considered aspects, further synthesized in Table~\ref{tab:axes1}, is given in the following paragraphs.

\paragraph{Scope.} The primary element adopted to organize the contributions is the main goal of the presented solutions, broadly aggregated into two main classes; analysis and hardening methods. Contributions devoted to the development of techniques and tools to evaluate the resilience of the application against hardware faults belong to the \textit{analysis methods} class, those that present new approaches to to enhance the capabilities of the system to detect and mitigate the effects of hardware faults are included in the \textit{hardening methods} class. Indeed some contributions introduce innovative strategies to evaluate resilience and exploit such information to tailor a hardening method; these methods are included in the \textit{both} class. Finally, some publications explore the application of either traditional or recent methods to specific study cases, reporting outcomes and limitations, experiences  others might benefit from; we classified them in the \textit{case studies} class.

\paragraph{Abstraction level.} Common to many fields of the digital systems' design area, approaches work at different levels of abstraction, within the entire hardware/software stack from the technological level to the application one. Moreover, multiple other aspects are highly dependent on the adopted abstraction level, therefore we prioritized it and identified the following six values, based on the main system element the proposed methods work on: 
\begin{itemize}
    \item \textit{Device} -- physical device,
    \item \textit{Logic} -- logic netlist.
    \item \textit{\ac{RTL}} -- architectural description at \ac{RTL} level,
    \item \textit{Microarchitecture} -- hardware schema described in the \ac{ISA},
    \item \textit{Algorithm} -- software elements within the implementation of the single \ac{DL} operators,
    \item \textit{Application} -- software elements in the dataflow graph of the \ac{DL} model.
\end{itemize}

\paragraph{Hardware platform.} The type of misbehavior caused by faults affecting the hardware in the application execution is highly dependent on the underlying platform. Therefore, another key aspect in the proposed analysis framework is the hardware platform where the \ac{DL} application is executed. Frequently adopted platforms are the \acp{GPU} and custom hardware accelerators, implemented on FPGA or ASIC; the CPU is used only in a few contributions, while the \ac{TPU} is increasingly receiving interest (\nuovo{e.g., the NVDLA platform~\cite{NVDLA}}).
As we will see, some contributions, especially when acting at the application abstraction level, will not consider any specific hardware, thus being platform independent or \textit{platform-agnostic}.

\paragraph{Fault model.} Every reliability study has a fundamental element driving the discussion, that is the source of the anomalous behavior the proposed approach is addressing. The reference abstraction level for the definition of the fault model is the logic/architecture one, where literature defines permanent models, such as the stuck-at faults, and transient ones, such as \ac{SEU}. Some of the proposed methods work at the application level, not referring to a specific hardware platform; it is therefore not possible to identify the mechanisms causing the anomaly in the expected values/behavior. For these contributions we added a \textit{functional} fault model, transient and/or permanent, according to the authors' specification. 

Since many of the analyzed works act at a higher abstraction level, fault models are generally abstracted to derive corresponding error models. 

\paragraph{Error model.} An error model describes the effects of the considered fault model at the selected abstraction level, and it affects one of the elements of the abstraction level. When working at device or RTL level, the relationship between fault and error are quite straightforward, when moving to higher abstraction levels, such a relationship is sometimes part of the contribution (for resilience analysis methods), sometimes omitted. Indeed, when adopting a functional fault model as previously discussed, fault and error models tend to be a unique element. Nevertheless, the error model is characterized by the specific corrupted \textit{location} which, once more, depends on the abstraction level.  
At device, \ac{RTL} and microarchitectural levels, fault locations typically include registers and memory elements storing processed data and the \ac{DL} model weights. At a higher levels of abstraction, error locations may also include parameters as single weights and bias constants, or data values, and even more complex data structures such as the outputs of the various neurons or the intermediate tensors produced by the layers in the \ac{DL} model. Therefore, we identify the following values:
\begin{inparaenum}[i)]
\item corrupted register/memory element,
\item corrupted parameter,
\item corrupted data value,
\item corrupted neuron output,
\item corrupted layer output.
\end{inparaenum}

\paragraph{\ac{ML} Framework.} The design of \ac{DL} applications is generally performed in specific \ac{ML} frameworks guiding and easing this type of activity by providing \ac{ML} operators already implemented and algorithms to automate the training and testing of the models. TensorFlow and PyTorch are examples of such frameworks. Several reliability studies and tools are developed and tailored for the specific \ac{ML} framework, to enable the integration of the resilience activity in the design flow and to exploit the elements it provides. This axis of the classification collects this aspect when specific to the proposed solution. 

\paragraph{Tool support.} The availability of open-source tools is indeed beneficial to the entire scientific community, to foster further developments as well as fair comparisons. Our framework includes also this aspect, to indicate whether the authors make available the developed software to perform the presented analysis/hardening solutions. The list of urls of the available software is reported in the last part of the paper.

\paragraph{Reproducibility.} Similarly to the previous aspect, we deemed relevant to be able to reproduce the outcomes of the study, in the future, to present a comparative analysis for supporting new solutions. To this end, we marked entries with a positive answer when the software is available or the adopted method is discussed in details allowing for it to be replicated.

\medskip Analysis and hardening approaches can be further characterized with respect to the specific proposed solutions, namely the dependability attribute, injection method and analysis output in the former approaches, target outcome, hardening technique and  hardening strategy in the latter. They are discussed in the following and summarized in Tables~\ref{tab:axes2} and~\ref{tab:axes3}, respectively.

\paragraph{Dependability attribute.} The various analysis approaches may focus on the evaluation of different attributes falling under the umbrella of the dependability; generally, works use to quantitatively analyze a reliability metric. In the considered scenario, further works analyze the vulnerability to faults of the various layers, operators or parameters composing the \ac{DL} model. Thus, the considered \textit{dependability attribute} is another characterizing aspect for the reviewed papers, that includes in our work the following values:
\begin{inparaenum}[i)]
\item reliability,
\item safety, and
\item vulnerability factor.
\end{inparaenum}

\paragraph{Injection method.} The vast majority of the analyzed contributions rely on fault/error injection methods to perform the resilience analysis, and the specific one depends on the abstraction level of the work. Here we list the following values to include all included studies: 
\begin{itemize}
\item Radiation tests -- the final system is irradiated with nuclear particles.
\item Fault emulation -- faults are emulated on the target hardware platform.
\item Error simulation -- processed data are corrupted during the execution of the software running non-necessarily on the target platform.
\end{itemize}

\paragraph{Analysis output.} When performing a resilience analysis, two main types of outcomes are typically reported: 
\begin{inparaenum}\item a quantitative measure adopted as a figure of merit, or \item a qualitative evaluation of the solution, based on various considerations.\end{inparaenum}
Sometimes, based on the analysis results, also guidelines for hardening the system are provided, often targeting the mitigation of the most susceptible elements in the analyzed \ac{DL} model. In the set of selected papers, all contributions on analysis methods report a quantitative output and eventually some hardening guidelines, that is what we report in the final synthesis.

\paragraph{Reliability property.} Hardening approaches can be classified w.r.t. the reliability property the final system will exhibit, that in the present set of studies is either fault detection or fault tolerance. 

\paragraph{Hardening technique.} In the \ac{DL} scenario, as in other contexts, often the the hardening process relies on redundancy-based techniques. Some approaches adopt the classical techniques, such as \ac{DWC} possibly coupled with re-execution, \ac{TMR}, \ac{NMR} and \ac{ECC}. Other works apply \ac{ABFT} or \ac{ABED} techniques within the single \ac{DL} operator, being the  algorithm generally based on matrix multiplications. Finally, a last class of works exploits specific characteristics of the \ac{DL} models, such as the adoption of fault-aware training strategies to exploit the intrinsic information redundancy in \ac{DL} models to deal with the effects of a fault.

\paragraph{Hardening strategy.} Finally, various strategies can be adopted aimed at reducing the overhead of hardening redundancies. In particular, apart from the application of a technique to the entire application, 
\textit{selective} hardening is used to protect only the most critical portion of the system and \textit{approximation} strategies can be used to limit the overheads of redundant application replicas. Finally, some solutions design \textit{specific} versions of \ac{DL} operators to obtain at their output a resilient result.

A detailed list of the collected values for each one of the framework axes is reported in Tables~\ref{tab:axes1} and~\ref{tab:axes2}. Indeed the framework can be extended in the future to include new relevant axes, and the values can always be incremented to cover newly reviewed solutions.

\begin{table*}[t]
\caption{Taxonomy axes}\label{tab:axes1}
\centering
\begin{tabular}{m{6em}>{\raggedright}m{15em}m{20em}}
\textbf{Classification Axis}    & \textbf{Description}   & \textbf{Values}   \\\hline\hline
\textit{Scope} & The focus of the approach  & Analysis (A), Hardening (H) or both (B)  \\\hline
\textit{Abstraction level}      & The abstraction level methodologies/solutions work at  & Device (DEV), Logic (LOG), \ac{RTL}, Microarchitectural (ISA), Algorithm (ALG), Application (APP)  \\\hline
\textit{Architectural platform} & The hardware where the application is executed. Affects the fault/error model, the abstraction level, etc. & CPU, GPU, TPU, FPGA, or any (in case of high abstraction-level methodologies)  \\\hline
\textit{Fault model} & The source of the anomalous behavior & \acf{SA}, \acf{SEU}, permanent functional (PFunc), transient functional (TFunc) \\\hline
\textit{Error model}      & The effects of the fault at the selected abstraction level, identifying the corrupted element & register/memory element (REG), parameter (P), data value (DV), neuron output (NO), layer output (LO)
\\\hline
\textit{\ac{ML} Framework}      & The exploited software \ac{ML} framework, if specified  & TensorFlow (TF, \cite{TensorFlow}), PyTorch (PT, \cite{IP+2021}), Keras (KE, \cite{GP2017}), Darknet (DK, \cite{DarkNet}), Caffe (CA, \cite{JS+2014}), TensorRT (TR, \cite{tensorRT}), cuDNN (cu~\cite{cuDNN}), N2D2 (ND~\cite{n2d2}), FINN (FI~\cite{UF+2017}), CMSIS-NN (CM~\cite{CMSIS}), CMix-NN (CN~\cite{CR+2020})  \\\hline
\textit{Tool support}      & Tools released  & Yes/No  \\\hline
\textit{Reproducibility}      & The possibility to replicate / compare against  & Yes/No \\
\hline\hline
\end{tabular}
\end{table*}

\begin{table*}[t]
\caption{Analysis studies: further classification}\label{tab:axes2}
\centering
\begin{tabular}{m{6em}m{15em}m{20em}}
\textbf{Classification Axis}    & \textbf{Description}   & \textbf{Values}   \\\hline\hline
\textit{Dependability attribute}    &  Attribute of interest & Reliability (Re), Availability (Av), Safety (Sa), Vulnerability factor (VF)  \\\hline
\textit{Injection method} & Fault injection method & Radiation (Ra), Emulation (Em), Simulation (Si), \nuovo{Analytical (An)}  \\\hline
\textit{Output} & Kind/type of output of the analysis & quantitative metrics (QM), guidelines for hardening (HG) \\\hline
\hline
\end{tabular}
\end{table*}

\begin{table*}[t]
\caption{Hardening studies: further classification}\label{tab:axes3}
\centering
\begin{tabular}{m{6em}>{\raggedright}m{15em}m{20em}}
\textbf{Classification Axis}    & \textbf{Description}   & \textbf{Values}   \\\hline\hline
\textit{Reliability property} &  Aim of the hardening & Fault detection (FD), Fault tolerance (FT)\\\hline 
\textit{Strategy} & Type of action & Full, Selective (Sel), Specific (Spec), Approximated (Ax) \\\hline
\textit{Technique} & Adopted technique & \acf{DWC}, \acf{TMR}, \acf{NMR}, \ac{DWC} + Re-Execution (D+R) \acf{ABFT}, \acf{ABED}, \acfp{ECC}, Checkpointing (CHK), \ac{DL}-specific (\ac{DL})\\\hline
\hline
\end{tabular}
\end{table*}

\section{The state of the art}\label{sec:soa}
We classified the reviewed papers primarily based on their main contribution partitioning them into \textit{analysis} methods and \textit{hardening} ones, those studies that work on both aspects have been included in the group associated with the predominant contribution. 

\begin{figure*}[t]
    \centering
    \includegraphics[width=\textwidth]{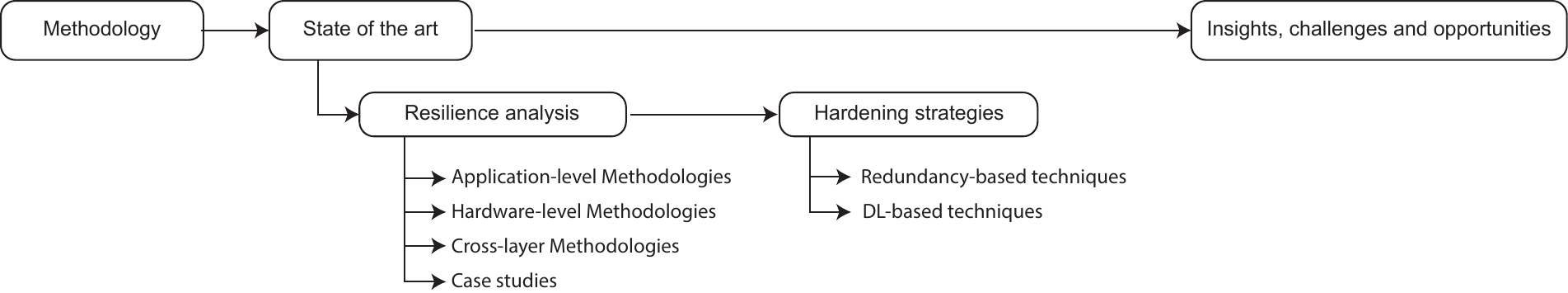}
    \caption{\nuovo{Paper organization.}}
    \label{fig:paperoutline}
\end{figure*}

\subsection{Resilience Analysis}

This first class of works includes approaches for the analysis of the resilience of digital systems running \ac{DL} applications w.r.t. the occurrence of faults. To further characterize them, we consider the abstraction level they work at, namely \textit{application-level}, \textit{hardware-level} or \textit{cross-layer}. 

Application-level methodologies aim at analysing the resilience of the \ac{DL} engine ignoring the underlying hardware platform. Therefore, such works consider the engine at the dataflow graph-level and study the impact of errors corrupting the weights of the model, the output of the operators or the variables within the operators' execution. The advantages of these methodologies are \begin{inparaenum}[i)] \item the possibility to apply them early in the design process, as soon as the \ac{DL} engine has been designed and trained; \item easiness of the deployment (no hardware prototypes and/or instrumentation is required); and \item the opportunity to work directly on the actual \ac{DL} engine that will then be used\end{inparaenum}. On the other hand, the solutions  may suffer from poor accuracy because of the abstract adopted error models. It is vital for application-level analyses to properly work that the adopted error models actually capture the effects that the faults in the hardware platform cause in the executed application otherwise inconsistent and only partially useful results are obtained.

Hardware-level methodologies exploit hardware-level fault injection platforms (mainly by emulating \acp{SEU} in the configuration memory of \acp{FPGA} or in the registers of \acp{GPU}) to accurately emulate the effects of faults in the hardware where the \ac{DL} model will be executed. These approaches are highly accurate because of the ability of reproducing the faulty behavior, and are time-wise more sustainable than simulation solutions, since fault injection can be executed at speed. On the other hand, these approaches are generally hard to be deployed, demanding specific hardware-level skills that a design team specialized in \ac{DL} may lack. Moreover, the application of resilience analyses belonging to this class are typically carried out late in the design process, thus making modifications expensive.

Finally, cross-layer methodologies try to bring together the advantages of the previous methodologies by splitting the analysis into two steps. First, a hardware platform-specific fault injection or radiation testing activity is performed on a portion of the \ac{DL} engine under analysis or on the single operators. In this way the actual effects of the faults occurring into an \ac{FPGA}, a \ac{GPU} or a CPU while accelerating/executing a \ac{DL} engine are captured. Then, the observed effects are used to feed a higher-level analysis/simulation engine to observe how these effects propagate through the subsequent layers of the model and if and how they affect the final output.

An additional group gathers a number of works that serve as case studies, because they apply to a specific \acp{DL} model, or actually report application case studies, presenting interesting results that are though specifically tailored for the discussed context. 
A brief description of contributions that belong to this class and to the above mentioned groups follows.

\subsubsection{Application-level Methodologies}

The paper in~\cite{BB+2019} presents one of the first tools for the resilience analysis of \acp{CNN} by performing error injection at application level. The tool, developed within the Darknet \ac{ML} framework, allows to corrupt the weights in the \ac{CNN} model and to carry out error simulation campaigns. The goal of the tool is to analyze the safety of the \ac{DL} applications; in particular, single experiments are classified as \textit{masked}, \textit{observed safe} and \textit{observed unsafe}; a threshold set to +/-5\%  is used to analyze the difference between the top ranked percentage in the erroneous result and the golden counterpart and to determine the safe/unsafe class the corrupted output belongs to.  
The paper considers permanent faults affecting the \ac{CNN} weights, not whatsoever relating these permanent functional faults to realistic faults in the underlying hardware running the application.

BinFI and TensorFI, presented in~\cite{CL+2019} and~\cite{NC+2022} respectively, are two subsequent contributions from the same research group, who, among other works, designed, developed and distributed two fault injection frameworks to evaluate \ac{ML} systems resilience. BinFI identifies safety-critical bits in \ac{ML} applications, while TensorFI analyzes the effects of hardware and software faults that occur during the execution of TensorFlow programs.  
The paper in~\cite{LR+2022} presents TensorFI+, an extension of the TensorFI environment. In particular, TensorFI+ supports TensorFlow 2 models, allowing to analyze also non-sequential models by corrupting the output of the layers. An interesting feature of the framework is the possibility to inject faults during the training phase of the \ac{CNN} under analysis.

PyTorchFI (presented in \cite{MA+2020}) is an error simulation engine for \acp{DNN} that exploits the PyTorch framework. The tool allows to emulate faults by injecting \textit{perturbations} in the weights and neurons of the convolutional layers of \acp{DNN}; the injected perturbations are functional errors, therefore no specific hardware architecture is considered. The analysis can be run on either CPUs or GPUs. 
A similar approach is implemented by Ares~\cite{RG+2018}, an application-level error simulator for \acp{DNN}\footnote{The tool is dated 2018, outside the boundary of this investigation. However, we included it, because it is adopted in several of the analyzed studies.}. Again, the tool supports the simulation of perturbations modeling faults affecting the weights, the activation functions and the state of the neurons. Several observations and guidelines are also drawn in the paper: \begin{inparaenum}[i)] \item the resilience of \acp{DNN} is strongly influenced by the data type and quantization of the weights; \item some classes are more likely to be mispredicted than others; \item faults in the weights are more likely to cause a misprediction than those in the activation functions; and, \item the more weights are reused the higher the failure probability.\end{inparaenum}

An analytical model called SERN is proposed in \cite{PTY2020} for the resilience analysis of \acp{CNN} w.r.t. soft errors affecting the weights. The results obtained by SERN are then validated against a set of fault injection experiments. In particular, by exploiting the proposed framework, the authors analyse the impact of the occurring faults w.r.t. \begin{inparaenum}[i)] \item the position of the affected bit within the stored value and \item the size of the stored value itself.\end{inparaenum} 
The authors further propose to harden the \ac{CNN} under analysis by protecting the most significant bits of the weights via \ac{ECC} and by selectively duplicating the first convolutions layers of the network.

\nuovo{The work in~\cite{RG+2023} addresses the problem of how to define a significant fault injection campaign. In particular, the paper presents a methodology for statistical fault injection aimed at sizing the fault injection campaign and selecting the most appropriate fault locations to achieve statistically significant results. The proposed method is specifically tailored to evaluate the weights of the \ac{CNN} models.}

Working at this abstraction level, the attention is focused on the performance and behavior of the \ac{DNN} with respect to different implementation strategies, when a fault corrupts its elements. More specifically, studies \cite{GS+2020,TL2021,SG+2019} explore the effects of quantization, compression and pruning on resilience. In particular, \cite{GS+2020} explores the impact of transient faults on compressed \acp{DNN} with respect to different pruning rates and data precision. The adopted fault model is the single bit flip on random live values stored in latches or registers and the authors develop a fault injection framework dubbed \textit{TorchFI} to emulate such effects. The main outcomes of this work are: \begin{inparaenum}[i)] \item 16-bit integer quantization can mitigate the overall error propagation w.r.t. the 32-bit floating-point baseline; \item while 16-bit quantization increases resilience, the more aggressive 8-bit quantization can produce a resilience drop; and \item pruned networks being smaller and faster will be less prone to faults, therefore possibly achieving a better resilience\end{inparaenum}. 
Similar quantization strategies are explored in \cite{TL2021}, proposing a simulator for evaluating the resilience of \acp{DNN} based on the frameworks of Keras and Tensorflow. The targeted fault model includes \acp{SEU} in the inputs, in the weights and in the output of the operators. 
Finally, the work presented in~\cite{SG+2019} discusses a simulation analysis for understanding the fault resilience of compressed \ac{DNN} models as compared to uncompressed ones. Simulation is then used to study the resilience of pruned and quantized \acp{DNN} w.r.t. not pruned and not quantized ones. The results presented in the paper demonstrate that on the one hand pruning does not impact the resilience of the \ac{DNN} while, on the other hand, data quantization largely increases it.

Another neural network element being tailored during the design and implementation of a system is the type of data, similarly to quantization. Approximation can be adopted to leverage model accuracy and implementation costs (e.g., execution time, hardware resource demand and power consumption). Since such representation choice has an impact on resilience, some studies investigate this aspect.
The authors in~\cite{RS+2021} exploit the application-level error simulator presented in~\cite{BB+2019} to analyze the safety w.r.t. the occurrence of permanent faults in the weights of two different \acp{CNN} when varying the data type; both floating point and fixed point data types at different precision levels are considered. 
The conclusions drawn in the paper are that the most resilient data type and precision level depend on the specific model; moreover, the paper suggests to select the best suited solution by trading safety and memory footprint of the various alternatives. \nuovo{Finally, the same authors have also analyzed in~\cite{GR+2023_2} the resilience of the novel POSIT data types, specifically defined for artificial intelligence computations, by means of the fault injection approach presented in~\cite{RG+2023}. Experimental results demonstrate how POSIT data types are less resilient than fixed point integer data types using a reduced precision.}

The authors of \cite{ZI+2022_1} use an ad-hoc designed \ac{ML} algorithm to build a \textit{vulnerability model} of the parameters of the \ac{DNN} under analysis. To reduce the number of required fault injection experiments to analyze the effects of bit flips, empirical considerations are introduced on the importance of the various bits within the value representation, both in the floating point and in the fixed point cases. The authors evaluate the benefits/loss of accuracy with respect to injecting faults in all locations showing that the outcome offers good opportunities.

\subsubsection{Hardware-level Methodologies}

Libano and others investigates in various studies the resilience of \acp{CNN} accelerated onto \acp{FPGA} by means of both radiations tests and fault emulation. In particular, in~\cite{LR+2021} radiation testing experiments are performed to analyze the impact of data precision and degree of parallelism on the resilience of the network. The conclusions of the study are: 
\begin{inparaenum}[i)] 
\item lower precision means less hardware resources and consequently lower fault probability; and 
\item more parallelism means more hardware resources but also faster execution thus, the best performance-resilience trade-off is reached with the highest achievable degree of parallelism.
\end{inparaenum}

An analysis of the effects of \acp{SEU} in \acp{BNN} accelerated onto SRAM-based FPGAs is presented in \cite{SP+2021}. The authors exploit the Xilinx FINN framework to build the \ac{BNN} and the FPGA Reliability Evaluation through JTAG (FREtZ) framework for the fault injection activity. The outcome of such logic-level fault injection experiment is subsequently exploited to carry out an in-depth layer-per-layer analysis of the effects of the faults on the accuracy of the network. The results of this study show that \acp{BNN} are inherently resilient to soft errors.

Additional examples of fault resilience analysis of \acp{CNN} accelerated onto \ac{FPGA} devices are presented in~\cite{WW+2021} and~\cite{XZ+2021}. In the former the authors explore alternative quantized designs and compare them against a classical \ac{TMR}, to evaluate costs and benefits. In the latter the authors consider permanent stuck-at faults and explore their effects, investigating four typical \acp{CNN}, including Yolo. The analysis shows that hardware faults can cause both system exceptions, such as system stall and abnormal runtime, and prediction accuracy loss. A custom evaluation metric based on accuracy loss is exploited, also taking into account system exception probability; the nominal and \ac{TMR}-protected versions are analyzed and compared against.

Another analysis for \acp{CNN} accelerated onto \ac{FPGA} devices is presented in \nuovo{\cite{GG+2023}}, where the focus is on investigating the impact of various pruning techniques on the resilience of the network. Several interesting considerations are drawn: \begin{inparaenum}[i)] \item removing filters that marginally contribute to the final classification increases the resilience of the \ac{CNN} w.r.t. fault in the configuration memory; \item networks with higher pruning rates are more robust to errors affecting the weights; and \item only a small percentage of weights (about 30\%) can (when corrupted) actually modify the behavior of the network and the percentage is even smaller if we consider the ability of causing an accuracy loss (about 14\%).
\end{inparaenum}
 \nuovo{The work in \cite{TI+2023} extends previous analyses by considering also BRAM to better focus on the various elements' susceptibility, to later apply a selective \ac{TMR}-based selective hardening strategy.}

A broad contribution to this class of solutions comes from Rech's team, analyzing the resilience to \acp{SEU} when executing \ac{DL} applications on \acp{GPU}. In particular, in~\cite{FdS+2019}, radiation tests are used to cause realistic \acp{SEU} in the target device; then, they complement the first set of experiments with microarchitectural-level fault injection by means of the SASSIFI tool, to collect a more extensive set of results. In the experiments, various versions of the same \ac{CNN} applications are analyzed, including the nominal versions and robust versions hardened by means of \acp{ECC} and \ac{ABFT} strategies applied to the convolutional layer. In a subsequent work~\cite{RM+2022}, the same research team evaluates with a similar approach the resilience of Google's \ac{TPU} by means of radiation testing. The most interesting aspect of this work is the definition of a set of error models in terms of the spatial patterns of the erroneous values in the output tensor of the convolution operator.

The work in \cite{CC+2021_1} presents a strategy to estimate the criticality of \acp{PE} in a systolic array with respect to faults that may permanently affect one of them, by building and training a \textit{neural twin}. The aim is to simplify the complexity (in terms of time) to analyze faults' effects with respect to solutions based on fault injection (as the authors did in the past) by using a trained model of the PE. The analysis on the single element offers the expected advantages and coherence with the PE real fault/error behavior, however the possibility to generalize and transfer the model to the rest of the \acp{PE} is still to be investigated. \nuovo{Finally, also the work in~\cite{PR+2023} focuses on the resilience of systolic arrays. In particular, the approach designs a \ac{RTL} simulator to inject stuck-at faults both in the weights and in the processed data and uses it to evaluate various architectural configurations w.r.t. the achieved performance and resilience. Experiments are executed on a very simple LeNet \ac{CNN}.}

\subsubsection{Cross-layer Methodologies}

Fidelity~\cite{HBL2020} is an accurate logic-level error simulator for \acp{DNN} accelerated via custom circuits. By exploiting a deep knowledge of the regular structure of \ac{DNN} hardware accelerators, Fidelity is able to reproduce and track in software the effects of \acp{SEU} occurring in the underlying hardware platform and affecting both the weights and the neurons. Moreover, based on the application of Fidelity to a set of large networks the authors draw the following considerations: 
\begin{inparaenum}[i)] 
\item not only the weights but also neurons and neuron scheduling highly affects the resilience of the network; 
\item the adopted data precision has an impact on the resilience; and, 
\item the larger the perturbation in the output of the neuron, the more likely the network suffers from a mis-classification.
\end{inparaenum} \nuovo{An evolution of the same method and tool has been later presented in \cite{HH+2023}, carried out on the NVDLA architecture~\cite{NVDLA}, used to investigate the effects of hardware faults (namely, single event upsets but applicable to other models) on training performance and accuracy. A detailed investigation is carried out, leading to valuable insights that can also be generalized to different platforms. Based on the outcomes, the authors propose a hardening solution based on a tailored partial re-execution of training runs when a problem is detected.} 

The work in \cite{LF+2021} presents an analysis framework aimed at predicting the propagation of \acp{SEU} affecting the registers of a CPU executing a \ac{CNN}. The SIMICS system simulator is employed to simulate the entire CPU and the executed \ac{CNN}; corruptions in the CPU registers are introduced to simulate \acp{SEU}. A small set of fault simulation experiments are first performed to extract data that are later used to train a Generative Adversarial Network (GAN). The GAN represents the actual core of the methodology since, after its training, it is used to predict, layer by layer, the percentage of faults that will be masked, those that will cause a crash and the ones that will lead to a \ac{SDC}.

The work in~\cite{MR+2021_1} presents another cross-layer error simulation framework; the approach is developed for a specific working scenario considering a microprocessor-based system running \acp{CNN}, focusing on faults affecting the RAM chip. The proposed approach is based on radiation experiments aimed at systematically analyzing the effects of the faults to build application-level error models, defined in terms of data corruption patters and occurrence frequencies; such models are specifically devoted to corrupt \ac{CNN} parameters, such as weights and bias constants. 
These models are integrated into an in-house error simulator offering the possibility to run \ac{CNN} resilience analysis at the application level, and, therefore, on any platform, without the need of actually deploying the \ac{CNN} on the target architecture.
The framework is used to evaluate the resilience of various implementations of the LeNet-5 \ac{CNN} obtained by using different data types, using different precisions.

A three-level resilience analysis environment is proposed in~\cite{CG+2022}. The first step is a profiling where each instruction of the \ac{DL} model under analysis is associated with information such as input values, output result and opcode by means of NVBit \cite{VS+2019}. As a second step, the microarchitectural fault injection for \acp{GPU} (called FlexGripPlus \cite{CD+2020}) is employed to characterize the effects of \acp{SEU} affecting the microarchitectural resources of the GPU cores while they are executing a single layer of the \ac{CNN}. Finally, the observed erroneous behaviors are used into a software-level fault simulation environment to analyse how faults propagate among the layers of the \ac{CNN}. This enables a detailed analysis of the vulnerability factor of every layer in the considered \ac{CNN}.

The work in~\cite{CC+2021} presents a cross-layer framework for the analysis of \ac{CNN} sensitivity against faults. The proposed framework relies on a CPU executing the \ac{CNN} and on an FPGA-based accelerator implementing the operator where faults have to be injected; the actual fault injection is realized by bit-flipping the content of the configuration memory of the FPGA device. 

CLASSES~\cite{BC+2022_2} is a cross-layer error simulation framework developed in the TensorFlow \ac{ML} framework. The tool is provided with a methodological approach to define error models starting from microarchitecture-level fault injection. More precisely, the method runs a preliminary fault injection campaign for each type of \ac{ML} operator on the target architectural platform; then, corrupted output tensors are analyzed to identify recurrent spatial patterns of erroneous values and their frequency. Thus, error models are defined for each one of these \ac{ML} operators in terms of an algorithmic description of how the output tensor of the operator should be modified according to the observed spatial patterns. Error models are stored in a repository used by the application-level error simulator that will run the entire \ac{CNN} model and will inject errors on selected intermediate tensors produced by any operator. Since the error model captures the effects of the fault corrupting the target architecture, error simulation is performed at application level, on any machine, without the need to deploy the application on the target final hardware. The paper demonstrates the effectiveness of the tool and the companion approach in the scenario of Yolo \ac{CNN} executed on a \ac{GPU}, however, the approach is general and can be employed for any architecture and \ac{CNN} model. \nuovo{The susceptibility to \acp{SEU} of the General Matrix Multiplication (GEMM), Fast Fourier Transform (FFT) and Winograd's convolution implementations has been studied in~\cite{BC+2023} by exploiting CLASSES. The authors first characterize the effects of the \acp{SEU} affecting the \ac{GPU} while executing the convolution operators; then, they analyze how the occurred faults impact on the overall \ac{CNN} accuracy. The remarkable outcome of the analysis is that the GEMM-based convolution is the most robust against \acp{SEU}.}

\nuovo{Similarly, SiFI-AI, presented in~\cite{HK+2023}, is a hybrid simulation environment that combines PyTorch inference with a cycle-accurate RTL simulator of \acp{SEU} in the registers of \acp{TPU}. saca-Fi~\cite{TW+2023} works at the same abstraction level and consists of an execution simulator, a fault injection module, and a reliability analysis framework to analyze both transient and permanent faults in the registers, providing an \ac{AVF} evaluation. Based on the outcome of the analysis, as case studies, the authors propose to harden the most sensitive registers and data parts by means of \ac{ECC} codes.}

\cite{DAS2022} presents FireNN; it is a cross-layer resilience analysis engine for \acp{CNN} accelerated onto FPGAs. The tool allows to study how the task carried out by the \ac{CNN} is affected by \acp{SEU} occurring either in the \ac{CNN} weights or in the layers output. More precisely, the entire \ac{CNN} is executed in software by means of the PyTorch framework while the \ac{CNN} operator cons the target for the fault injection in transferred onto the FPGA device. Once the operator has been configured in the FPGA the fault is injected, the (possibly corrupted) operator output is collected and it is then reintroduced in the subsequent operators that, again, are executed in software. 

LLTFI~\cite{ACP2022} supports framework-agnostic fault injection in both C/C++ programs and \ac{ML} applications written using any high-level \ac{ML} framework. It uses LLVM to compile the \ac{DNN} model in the \ac{IR} targeted for CPU platform, that is used for fault injection activities. In this way, the tool supports injection at the granularity of single \ac{IR} instructions, allowing also to observe at a fine-grain level the error propagation among the various parts of the \ac{DNN}. Based on these capabilities, LLTFI provides guidelines and metrics to drive the selective instruction hardening, as demonstrated by the experimental activities discussed in the paper.

\nuovo{A framework, called DeepAxe, for the analysis and the design space exploration of the effects of approximation and the trade-off between area occupation and reliability is presented in~\cite{TR+2023}. In particular, DeepAxe targets \ac{DNN} accelerators implemented onto \acp{FPGA}. The framework starts with a Keras description of the \ac{DNN} that is used to measure the \textit{ground-truth} accuracy. Then, the Keras model is translated in C and fault simulation is performed to measure the reliability of the model. Finally, high-level synthesis is applied to obtain the hardware description and approximation is applied thus allowing to evaluate the area occupation of the final circuit.}

\subsubsection{Case studies}

The paper in \cite{LH+2019} from NVIDIA analyses the reliability and safety of a \ac{CNN} (executed on a \ac{GPU}) for object detection in the automotive application domain. Both fault simulation and radiation testing are carried out. It is one of the few papers where safety issues (Failure in Time in particular) are taken into account. The paper highlights how the use of \acp{ECC} for the protection of the content of the memory of the \ac{GPU} increases the reliability of the system. On the other hand, the paper also states that \ac{ECC} protection is not enough and that periodic structural tests are recommended to mitigate risks due to \acp{SEU}.

The impact of \acp{SEU} occurring in the weights on the accuracy of \acp{CNN} is analysed in \cite{NA+2019} via an ad-hoc designed fault simulation framework. GoogleNet, Alexnet, VGG16, and SqueezeNet are considered in the analysis and the target hardware platform is a \ac{GPU}. The analysis is carried out targeting three aspects: data representation (fixed point versus floating point values), position of the corrupted bit within the value and position of the corrupted layer within the network. The outcome of the analysis refers that \begin{inparaenum}[i)] \item \acp{CNN} using fixed point values are much more resilient than the ones using floating point values; and, as expected, \item faults occurring in the exponent of floating point \acp{CNN} have the biggest impact on resilience; and \item the last layer of the network are the ones having the biggest impact on its resilience.\end{inparaenum} 

The works in \cite{IWB+2019,IWA2020} deal with two different case studies, analyzing and improving the resilience of ResNet and GoogLeNet implemented on \acp{GPU}, respectively. In both cases the context is very specific such that, as the authors state, it is not possible to generalize the outcomes that, thus, can actually be exploited only in similar application contexts. Layer and kernel vulnerability is analyzed by performing a fault injection campaign via SASSIFI, to identify the most vulnerable aspects of the implemented model. 
In \cite{IWB+2019} the authors also selectively harden some of the kernels that exhibited high vulnerability, by triplicating them and voting the output.

The paper in~\cite{MR+2021} presents an analysis of the resilience against \acp{SEU} affecting the weights of the LeNet5 \ac{CNN} applied to the MNIST dataset. Based on the results of this analysis the authors draw several considerations: \begin{inparaenum}[i)] \item faults affecting the convolutional layers are more likely to cause a significant accuracy drop than faults affecting the fully connected layers; \item the faults affecting the exponent of the floating point values used to represent the weights have the largest effect on the accuracy of the \ac{CNN}; \item the use of Sigmoid operators instead of ReLU ones decreases the resilience of the \ac{CNN}; and \item average pooling is more capable of preventing the propagation of faults compared to max pooling.\end{inparaenum}

\nuovo{In~\cite{KB+2021} the reliability and safety analysis of a systolic array against stuck-at faults occurring in the datapath, i.e., weights, bias, multiplier, and accumulator units, is presented. The analysed faults are classified based on the severity of the effects they cause on the output of the systolic array. Based on such analysis, the paper additionally presents two algorithms for test pattern generation meant to detect the most critical faults. Similarly, a simulation analysis of the effects of permanent faults in the datapath occurring during the training of \acp{TPU} has been presented in~\cite{HL2023}. The authors observe three possible effects: faults that are totally masked, faults that cause NaN/Inf values and faults that cause a sharp accuracy degradation. Finally a simple fault detection and reaction scheme is proposed: a training iteration is discovered to have suffered from a fault as soon as the training loss value exceeds a pre-configured bound; then, the two most recent training iterations are re-executed in order to recover from the fault.}

\subsection{Hardening Strategies}
The second class of reviewed works propose approaches for the hardening of systems running \ac{DL} applications w.r.t. effects of faults corrupting the underlying hardware. These works focus on handling and mitigating \acp{SDC}, constituting the most dangerous effect of faults, because it is not detected by the system; a few contributions deal also with the recovery from \ac{DUE}.
This class of works can be further partitioned into  
\begin{inparaenum}
\item approaches applying classical redundancy-based hardening strategies, and
\item design strategies exploiting peculiar characteristics of \ac{DL} models.    
\end{inparaenum}
One of the main challenges in the hardening process is the fact that \ac{DL} applications are compute intensive; therefore selective or approximated techniques are generally defined when considering redundancy-based strategies, to limit the costs of the hardening process.
Moreover, \ac{DL} models are internally redundant and presents specific peculiarities that can be exploited to introduce a degree of intrinsic resilience to faults in the designed applications. Thie second group of works exploits these properties to define resilience-driven design methods.

\subsubsection{Redundancy-based techniques}
The work in~\cite{MH+2021} proposes two complementary selective hardening techniques for introducing fault tolerance in \ac{DL} systems acting at application level, without targeting any specific architecture. The first technique works at design time to identify the most vulnerable feature maps. This vulnerability analysis is performed by means of metrics to estimate 
\begin{inparaenum}[i)]
\item the probability of activation of a fault while processing a feature map and
\item the probability of propagation of the generated error to the primary outputs of the \ac{CNN}.
\end{inparaenum}
Then, most vulnerable feature maps are hardened by means of \ac{DWC}, and, in the case of mismatch, re-execution is performed at run-time.
The second proposed technique works at run-time and monitors with an \ac{ABED} approach the outputs of each \ac{CNN} inference. In particular, two metrics are used to classify the outputs as \textit{suspicious}, and if needed, a re-execution is performed to recompute the results. These two metrics are defined based on empirical observations showing that, when considering a \ac{CNN} for classification activities, the difference between the top two confidence classes exhibits a strong inverse relationship with the occurrence of a misclassification. The extensive experimental evaluation of the proposed techniques is performed in PyTorchFI, by the same authors, and is architecture agnostic.

The work in~\cite{SHT2021} proposes a hardening approach based on selective application of classical redundancy-based techniques against both transient faults in computations and permanent faults in the memory storing the weights. The approach exploits techniques for explainable \ac{AI} to identify the most susceptible locations in the \ac{CNN} at the granularity of the single weight, and neurons in the feature map whose corruption will possibly cause a misclassifications with a high probability. Then, \acp{ECC} and \ac{TMR} are selectively applied to the most critical weights and neurons, respectively. Even if the approach works at the application level and is prototyped in the PyTorch framework, it is particularly tailored for \acp{DNN} designed by using a low data precision, generally accelerated in hardware. 

The authors in~\cite{GG2021} develop a so-called \textit{Resilient TensorFlow} framework, obtained by adding to TensorFlow a set of fault-aware implementations of its base operators, to address \acp{SEU} occurring in the underlying \ac{GPU} device. Each new operator is implemented to execute a thread-level \ac{TMR}ed version of the nominal counterpart. Then, thread blocks are opportunistically scheduled and distributed on the \ac{GPU} cores to avoid a single fault to corrupt multiple redundant threads.
The proposed approach is validated by means of both application-level fault simulation, by means of TensorFI, and microarchitectural-level fault emulation, by means of NVBitFI~\cite{TH+2021}. An interesting further contribution is the introduction of the Operation Vulnerability Factor, a metric used to evaluate the resilience of operations, to validate the proposed solution. In our opinion, the metric could be adopted to compare different solutions focused on hardening the single operator.

The work in~\cite{AG+2021} puts together various preliminary contributions previously published by the same research group to harden \acp{CNN} executed on ARM CPUs. In particular, they evaluate through simulated fault injection at microarchitectural level, by means of the SOFIA tool~\cite{GB+2022_2}, the resilience of various implementations of the same \ac{CNN} with various data precision models (integers at 2, 4, and 8 bits). Based on the results, they harden the various \acp{CNN} by using two different techniques: 
\begin{inparaenum}[i)]
\item a partial \ac{TMR} applied at instruction level on sub-parts of the application, or
\item an ad-hoc allocation of variables to registers. 
\end{inparaenum}
The idea at the basis of this second technique is that minimizing the number of used memory elements reduces the area exposed to radiations and therefore system resilience improves, here measured in terms of Mean Work To Failure (MWTF). The experimental analysis is performed on the MobileNet \ac{CNN}.

The work in~\cite{BC+2022_1} proposes a selective hardening approach for \acp{CNN}. First, the approach uses the CLASSES error simulator~\cite{BC+2022_2} to characterize the vulnerability against \acp{SEU} of each layer in the \ac{CNN}. This metric is defined as the percentage of faults corrupting the single layer causing the final \ac{CNN} output to be functionally different from the golden one, i.e., \textit{unusable} as defined in~\cite{BB+2022}. As an example, when considering an image classification task, the output of the \ac{CNN} is \textit{usable} when the input image is correctly classified, even if the actual output percentage values are slightly different from the golden ones; on the other hand, the output is \textit{unusable} when the output percentage values are highly corrupted thus causing a misclassification of the input image. Then, the overall robustness of the \ac{CNN} is computed by combining the layers' vulnerability factors. The approach performs an optimization of the hardening based on a selective layer duplication to co-optimize the overall robustness of the \ac{CNN} and its overall execution time. The approach is applied to a set of 4 different \ac{CNN} applications targeting a \ac{GPU} device.

Another example of application-level selective hardening approach is the strategy in~\cite{RG+2022}, that exploits a resilience score previously defined in~\cite{RS2021} to rank neurons in the model; then, the approach prunes neurons classified as non critical to reduce memory footprint, and triplicates neurons classified as critical to improve model resilience. The strategy is implemented in the PyTorch framework without targeting a specific hardware platform, and the resilience of the system is evaluated against errors randomly modifying or setting to zero the output values of the single neurons.

SHIELDeNN~\cite{KR+2020} and STMR~\cite{BG+2022} are two similar approaches, targeting \acp{BNN} implemented on \acp{FPGA}. Both tools perform a preliminary vulnerability analysis of the parameters of the \ac{BNN} (in particular, weights and activation functions) to identify the most critical ones; this analysis is based on in-house fault simulators. Then, selective \ac{TMR} is applied to the most critical parameters, at the granularity of entire layers in~\cite{KR+2020} and individual channels in~\cite{BG+2022}. Although both works target FPGA devices, they only harden against faults affecting the data memory storing \ac{BNN} parameters, neglecting faults affecting the device configuration memory, whose corruption actually leads to a modified functionality.

Still targeting FPGAs, \cite{GZ+2022} presents a methodology for achieving a lightweight fault tolerancefor \acp{CNN}. The idea is to avoid the classical \ac{TMR} scheme by adopting an approximated \ac{NMR}-based approach; instead of having three exact replicas of the \ac{CNN} plus a voter, the proposed methodology exploits the so-called \textit{ensemble learning}, an approach used in \ac{DL} for increasing model accuracy. In particular, the technique introduces a number of redundant \acp{CNN}, that are simpler and smaller than the original one. During the training phase each \ac{CNN} learns a \textit{subset} of the problem; then, during testing/deployment all \ac{CNN} output responses are \textit{merged} by a \textit{combiner} module that produces the final output as the original \ac{CNN} would have computed. The methodology is applied to various versions of the ResNet \ac{CNN} and the resilience evaluation is performed by means of a fault injector corrupting the \ac{FPGA} configuration memory.

The work in~\cite{DW+2021} targets a hardware accelerator organized as a dataflow architecture for \ac{ML} acceleration.
The strategy exploits computing elements in the architecture currently having as activation value a zero or an identical value of a neighbor computing element; the aim is to duplicate the same computation of the neighbor element. Additional logic is introduced into the architecture to manage on-the-fly duplication of the computations, to check results and, if needed, to re-execute faulty elaborations. The advantage of the approach is to benefit from the massively parallel nature of the considered \ac{ML} accelerator to introduce computation replicas at execution level without extending the architecture with additional computing elements. The architecture is experimentally validated onto an FPGA device by performing emulated fault injection in the registers of the \ac{RTL} description.

\cite{ZD+2021} introduces several \ac{ABFT} schemes to detect and correct errors in the convolutional layers during the inference process; to this end the authors develop in Caffe framework a \textit{soft error detection library for \acp{CNN}}, \textit{FT-Caffe}. The approach is based on the adoption of checksum schemes and layer-wise optimizations, opportunely calibrated by means of a workflow that provides error detection and then error correction. Being it a runtime method, performance degradation is overhead aspect traded-off against fault resilience. Application-level error simulation is used by means of an in-house tool to evaluate the approach.

Two \ac{ABED} techniques are proposed in~\cite{HS+2022} and~\cite{KR2021} for linear layers, i.e., convolutional and fully-connected layers. Both works, targeting \ac{GPU} devices, are based on computation and checksum validation in matrix multiplication algorithms. In particular, the approach in~\cite{HS+2022} considers quantized models and is implemented in CUDA, using also the cuDNN library; the other layers are protected by traditional \ac{DWC}. The experimental evaluation is performed through microarchitectural-level fault emulation by injecting single bit-flips in the layer inputs and outputs and weights, and through radiation testing.
The approach in~\cite{KR2021} defines two different checksum strategies: 
\begin{inparaenum}
\item a global one, being a refined version of the classical hardening scheme for matrix multiplication, and
\item a thread-level one, where the classical scheme is redesigned to aggressively use the \ac{GPU} tensor cores.
\end{inparaenum}
A design-time profiling approach, called \textit{intensity-guided \ac{ABFT}}, is used to decide for each \ac{CNN} layer which strategy is the most efficient one in terms of execution time. The paper presents only an experimental evaluation of the performance of the proposed approach, neglecting reliability measures.


It is worth mentioning another similar \ac{ABED} strategy based on checksums~\cite{OO2020_1}, applicable to convolutional and fully-connected layers. As for the previous contributions, the authors propose a hardware module to accelerate computation and checksum validation. The evaluation is again performed at the application level within a custom error simulation environment developed in Keras and Tensorflow.

\nuovo{A hardening methodology based on a selective \ac{ECC} application, dubbed harDNNing, is presented in~\cite{TKS2023}. The framework first performs fault injection experiments in the parameters of the various layers of the \ac{DNN} model. As a second step, based on the results of these fault injections, \ac{ML} models are trained to predict the criticality of all the parameters and of all the bits within a single parameter. Finally, \ac{ECC} are selectively inserted to protect the previously identified critical bits and critical parameters, thus achieving low-overhead fault tolerance.}

\nuovo{An analytical model to study the propagation of \acp{SEU} affecting the weights of a \ac{CNN} has been proposed in~\cite{YS+2020}. In particular the authors define the concept of \textit{SEU-Induced Parameter Perturbation (SIPP)} as the modification of the value of a \ac{CNN} weight caused by an \ac{SEU}. Once the possible SIPPs have been identified, the authors study if and how they propagate to the output of the \ac{CNN} and based on this analysis the most critical weights are identified. As a final step, \ac{TMR} or \ac{ECC} are applied to the most critical weights to increase the robustness of the \ac{CNN}.}

\subsubsection{\ac{DL}-based techniques}

The paper~\cite{CLP2021} introduces \textit{Ranger}, a fault correction technique identifying and modifying values presenting a deviation from the nominal ones, presumably due to the occurrence of transient faults in the processed data. The intuition at the basis of this technique, previously discussed in the paper presenting BinFI~\cite{CL+2019}, is that each layer in a \ac{DNN} model produces in output tensors containing elements included in a specific value range. Moreover, if a \ac{SEU} generates a corrupted value in the output tensor sensibly different from its nominal range, there is a high probability that this will cause the \ac{DNN} to generate an erroneous output, an event that does not occur when the corrupted values is anyway within the nominal range. Thus, the proposed low-cost technique consists in introducing on the output of selected \ac{DNN} layers a new operator that clips those output values that are outside identified restriction bounds. The proposed idea is implemented in TensorFlow and evaluated by means of TensorFI. 

Paper~\cite{HHS2020} presents a technique very similar to Ranger. The paper considers permanent faults in the weights of the \ac{DNN} and defines a novel clipped version of the ReLU activation function, replacing output values larger than a given threshold with a 0. A methodology is proposed to identify a proper threshold capable of identifying possible faults causing out-of-range corrupted values and at the same time limiting the negative impact of this new operator on the accuracy of the overall \ac{DNN}. The experimental evaluation is carried out by means of an in-house error simulator developed in PyTorch.

The work in~\cite{GS+2022} proposes yet another value range limiting strategy, implemented by modifying the activation function to perform a clipping against a threshold. 
Based on the limitations of previous efforts in the same direction, the authors employ a fine-grained neuron-wise activation function, to be determined in a supplementary training phase, that follows the traditional accuracy training. To this end, the work proposes a two-steps framework that supports the design and implementation of a resilient \ac{DNN}. The authors analyze the final implementation against memory faults, that is weights and biases of different layers, as well as parameters of activation functions. An in-house error simulator is developed in PyTorch for running an experimental evaluation. Results are compared against hardening solutions proposed in~\cite{HHS2020} and~\cite{CLP2021}, showing an improvement.

Few other papers present alternative strategies to address faults causing high-magnitude errors. For instance, the work in \cite{OO2021} combines quantization tailored on the parameter distribution at each \ac{DNN} layer and a training method considering a specific loss function, optimistically exploiting the selected quantization scheme not to decrease the accuracy while pursuing a high resilience. This approach, validated in an ad-hoc application level error simulation framework developed in PyTorch, outperforms two different strategies proposed by the same authors 
and a state-of-the-art approach based on explicit value range clipping~\cite{HHS2020}. Another work exploiting the statistical distribution of the tensor values is proposed in~\cite{AM+2022}; it defines thresholds for localizing and suppressing errors. The technique is coupled with state-of-the-art checksum strategies for error detection. 
The authors in~\cite{GS+2022_1} also exploit the statistical distribution of the values in the output of the \ac{DNN}, before applying the final softmax normalization, to detect outliers, which represent a suspicious symptom of a fault corrupting the system.

In this class of papers, we found papers that optimize the memory overhead introduced by the application of  \ac{ECC} to the \ac{DNN} weights by exploiting peculiar properties and characteristics of \ac{DNN} models. 
As an example, the study in~\cite{GN+2019} proposes a novel training scheme, namely Weight Distribution Oriented Training (WOT), to regularize the weight distribution of \acp{CNN} so that they become more amenable for protection by encoding without incurring in overheads.
The idea is to exploit the fact that weights in a well-trained \ac{CNN} are small number, requiring a few bits to be represented with respect to the available ones. Therefore, part of the bits are used to hold the \ac{ECC}, effectively using a 8-bit quantization strategy for the weights, to use the remaining bits for the checksum. The evaluations is performed at application level by means of a custom fault simulation method in PyTorch. Another similar work is presented in~\cite{LY2022_1} where a Double Error Correcting code based on parity is adopted to protect weights against stuck-at faults. The proposed approach, prototyped in Keras, outperforms the one in~\cite{GN+2019}.

Finally, other papers follow the same path, also broadening the field of analysis. As an example the authors in~\cite{BEA2021} continue the analysis of the robustness of the various data types by considering the recently introduced \textit{Brain-Float 16 (bf16)} format; since this data type is obtained by removing 16 bits from the mantissa of the standard 32 bit floating point, it presents a higher vulnerability to faults. Based on the robustness analysis, the authors define another similar coding scheme for the weights of the model. In particular, to avoid any memory overhead, a parity code is applied by using the \ac{LSB} of each word as the checking bit; the intuition is that a change in the \ac{LSB} marginally affects the model accuracy. Then, when a parity error is detected, the entire weight is set to zero; in fact, as studied in~\cite{ZG+2018}, a change of a single weight to zero generally does not affect the \ac{DNN} result.

A novel hardening paradigm, dubbed \textit{fault-aware training} is proposed in~\cite{ZG+2020, CS+2022}. The idea behind this technique is to inject faults during the training process thus forcing the \ac{CNN} to learn how to deal with the occurrence of faults at inference time. This promising technique on the one hand enables a \textit{low-cost} hardening but on the other hand it poses new challenges to the designer. Indeed, it is vital to identify the proper amount of faults to be presented to the \ac{CNN} during the training phase; a high number could increase robustness, introducing the side-effect of preventing training convergence and producing an excessively large \ac{CNN}. A reduced number of faults will result in a quick but possibly ineffective training. 
In the paper the newly proposed fault-aware training is coupled with two additional \ac{CNN} model modifications aimed at mitigating high-magnitude errors:
\begin{inparaenum}[i)] 
\item replacing the standard ReLU activation with its clipped counterpart, ReLU6 (originally proposed in \cite{SH+2018}); and 
\item re-ordering the layers in the \ac{CNN} such that ReLU6 is always executed before batch normalization.\end{inparaenum}
The paper evaluates the proposed approach by considering a \ac{GPU} target device and by using both microarchitectural fault injection (via NVbitFI) and application level error simulation (via a Python-based in-house tool).
Fault-aware training is also investigated in~\cite{BS+2022}, where the authors introduce specific loss functions and training algorithm to deal with multiple bit errors. The evaluation is carried out at the application level by not considering any specific hardware platform.

\textit{Fault-aware weight re-tuning} for fault mitigation is proposed in~\cite{SH2023}. In this paper the authors first analyze the resilience against permanent faults of a \ac{MAC} structure generally used in \acp{GPU} and \acp{TPU}. In particular, the authors analyse how the structure is sensitive to \ac{SA} faults a \ac{CNN} is w.r.t. 
\begin{inparaenum}[i)] 
\item the degree of approximation adopted in the employed multipliers; 
\item the position of the faulty bit in the corrupted value; and \item the position of the layer affected by the fault in the whole \ac{CNN}.\end{inparaenum} 
The authors propose to prune the weights that are mapped on the corrupted bits and that are thus going to be affected by the \ac{SA} faults (previously identified through post-production test procedures). Once such pruning has been carried out, re-training of the \ac{CNN} is performed. The experimental evaluation is performed by designing a systolic array architecture based on the considered \ac{MAC} structure. Fault injection campaigns are run with an in-house error simulator in TensorFlow.

The work in~\cite{LL+2021} first performs a systematic analysis of the \ac{PVF} of the various instructions of an ARM CPU executing \ac{DL} applications. Experiments are performed by means of a fault emulation tool corrupting the \ac{ISA} registers by means of the on-chip debugging interface. Then, it defines two techniques to harden the considered system against \acp{SDC}: 
\begin{inparaenum}
\item selective kernel-level \ac{DWC} with re-execution, and 
\item a \textit{symptom-based} technique checking all values of the intermediate results against a given threshold to trigger a re-execution when a value is above it.
\end{inparaenum}
This second technique is based on the same intuition of the range restriction strategies discussed above (e.g.,~\cite{HHS2020,CLP2021}).
Finally, the paper considers the adoption of kernel-level check-pointing to recover from crashes or other \ac{DUE}.
In a subsequent work~\cite{LY2022_1}, the same authors note that output values of a \ac{DNN} layer present a regular data distribution that can be analyzed at runtime to compute, during the inference process, the two thresholds to be used for the range restriction technique.

\cite{ZH+2019} focuses on a different perspective with respect to all previous contributions: the impact of faults during model training. An in-house error simulator is defined within the Caffe framework to inject bit-flips in the variables to simulate \acp{SEU} affecting the \ac{HPC} system running the training procedure. Outcomes of such an analysis are that (as already emerged in other works for errors affecting floating point values and layers)   
\begin{inparaenum}[i)]
\item most training failures result from higher order bit flipping in the exponents, and
\item convolutional layers are more failure prone.
\end{inparaenum}
Moreover, the authors highlight how monitoring the value of the loss function among the various training iterations is an effective signal to detect most of the \acp{SDC} causing a training failure. Based on this observation, an ad-hoc error detection strategy is defined for training failures due to \acp{SEU}.

\nuovo{A mitigation methodology without redundant hardware and without model retraining for permanent faults is systolic arrays is presented in~\cite{HY+2023}. The method exploits fault maps generated during post-fabrication testing to arrange significant data to \acp{MAC} with fewer faults. Moreover, the authors propose to compensate the effect of a fault by \textit{correcting} the faulty value substituting it with the value of the sign (they call this technique \textit{sign compensation}).}

\begin{table*}[t]
\caption{Contributions according to their type.\label{tab:category}}
\centering
\begin{tabular}{ll}
Resilience analysis & \\
\phantom{space}Application-level methodologies & \cite{RG+2018}\cite{BB+2019}\cite{CL+2019}\cite{NC+2022}\cite{LR+2022}\cite{MA+2020}\cite{PTY2020}\nuovo{\cite{RG+2023}}\cite{GS+2020}\cite{TL2021}\cite{SG+2019}\cite{RS+2021}\nuovo{\cite{GR+2023_2}}\cite{ZI+2022_1}\\
\phantom{space}Hardware-level methodologies & \cite{LR+2021}\cite{SP+2021}\cite{WW+2021}\cite{XZ+2021}\nuovo{\cite{GG+2023}}\nuovo{\cite{TI+2023}}\cite{FdS+2019}\cite{RM+2022}\cite{CC+2021_1}\nuovo{\cite{PR+2023}}\\
\phantom{space}Cross-level methodologies & \cite{HBL2020}\nuovo{\cite{HH+2023}}\cite{LF+2021}\cite{MR+2021_1}\cite{CG+2022}\cite{CC+2021}\cite{BC+2022_2}\nuovo{\cite{BC+2023}}\nuovo{\cite{HK+2023}}\nuovo{\cite{TW+2023}}\cite{DAS2022}\cite{ACP2022}\nuovo{\cite{TR+2023}}\\
\phantom{space}Case studies & \cite{LH+2019}\cite{NA+2019}\cite{IWB+2019}\cite{IWA2020}\cite{MR+2021}\nuovo{\cite{KB+2021}}\nuovo{\cite{HL2023}}\\
Hardening strategies & \\
\phantom{space}Redundancy-based techniques & \cite{MH+2021}\cite{SHT2021}\cite{GG2021}\cite{AG+2021}\cite{BC+2022_1}\cite{RG+2022}\cite{KR+2020}\cite{BG+2022}\cite{GZ+2022}\cite{DW+2021}\cite{ZD+2021}\cite{HS+2022}\cite{KR2021}
\cite{OO2020_1}\nuovo{\cite{TKS2023}}\nuovo{\cite{YS+2020}}\\
\phantom{space}\acl{DL}-based techniques & \cite{CLP2021}\cite{HHS2020}\cite{GS+2022}\cite{OO2021}\cite{AM+2022}\cite{GS+2022_1}\cite{GN+2019}\cite{LY2022_1}\cite{BEA2021}\nuovo{\cite{ZG+2020}}\cite{CS+2022}\cite{BS+2022}\cite{SH2023}\cite{LL+2021}\cite{ZH+2019}\nuovo{\cite{HY+2023}}\\
\end{tabular}
\end{table*}


\vskip 1em

The adoption of the two identified main classes, namely \textit{resilience analysis} and \textit{hardening strategies}, to partition the reviewed contributions allows us to organize them based on the main focus of the novelty of the presented solution. Table~\ref{tab:category} offers a bird's-eye view of this classification and summarizes the outcome.

As mentioned, the classification framework we define allows us to capture the elements we deem more relevant emerging from the reviewed contribution, thus providing a guide in identifying pertinent state-of-the-art proposals to build upon or to compare against. Table~\ref{tab:all} collects the \numintabs{} entries of the analyzed papers for an easy access to the information.

\onecolumn

\begin{landscape}
\begin{longtable}{|l|l|lllllll|lll|lll|}
\caption{Contribution Classification}\label{tab:all} \\

\hline
\textbf{} & \textbf{} & \textbf{} & \textbf{} & \textbf{} & \textbf{} & \textbf{} & \textbf{} & \textbf{} 
& \multicolumn{3}{c|}{\textbf{Analysis}} & \multicolumn{3}{c|}{\textbf{Hardening}} \\
\begin{sideways}\textbf{Paper}\end{sideways} & \begin{sideways}\textbf{Scope}\end{sideways} & \textbf{Abs.} & \textbf{HW} & \textbf{Fault} 
& \textbf{Error} & \textbf{\ac{ML}}  & \textbf{Tool} & \textbf{Rep.}  
& \textbf{Dep.} & \textbf{Inject} & \textbf{Out} & \textbf{Rel.} & \textbf{Tech.} & \textbf{Str.} \\    
\textbf{} & \textbf{} 
& \textbf{Lev.} & \textbf{Plat.} & \textbf{Model} 
& \textbf{Model} & \textbf{Fram.}  & \textbf{} & \textbf{} 
& \textbf{Attr.} & \textbf{Meth.} & \textbf{} 
& \textbf{Pro.} & \textbf{} & \textbf{}  \\

\endfirsthead

\multicolumn{15}{c}%
{{\bfseries \tablename\ \thetable{} -- continued from previous page}} \\
\hline
\textbf{} & \textbf{} & \textbf{} & \textbf{} & \textbf{} & \textbf{} & \textbf{} & \textbf{} & \textbf{} 
& \multicolumn{3}{c|}{\textbf{Analysis}} & \multicolumn{3}{c|}{\textbf{Hardening}} \\
\begin{sideways}\textbf{Paper}\end{sideways} & \begin{sideways}\textbf{Scope}\end{sideways} & \textbf{Abs.} & \textbf{HW} & \textbf{Fault} 
& \textbf{Error} & \textbf{\ac{ML}}  & \textbf{Tool} & \textbf{Rep.}  
& \textbf{Dep.} & \textbf{Inject} & \textbf{Out} 
& \textbf{Rel.} & \textbf{Str.} & \textbf{Tech.} \\    
\textbf{} & \textbf{} 
& \textbf{Lev.} & \textbf{Plat.} & \textbf{Model} 
& \textbf{Model} & \textbf{Fram.}  & \textbf{} & \textbf{} 
& \textbf{Prop.} & \textbf{Meth.} & \textbf{} 
& \textbf{Out.} & \textbf{} & \textbf{}  \\

\hline
\endhead

\hline\multicolumn{15}{r}{{Continued on next page}} \\
\endfoot

\hline \hline
\endlastfoot

\hline\hline

\cite{RG+2018} 
& A 
& APP & any & PFunc/TFunc
& P/DV/NO & KE & Yes & Yes
& Re & Si & \phantom{}\footnote{QM is always present and therefore omitted, we only report HG when applicable.}
& - 
& - & - \\

\cite{BB+2019} 
& A 
& APP & any & PFunc
& P & DK & No &
Yes &  
Sa & Si & HG 
& - & - & - \\

\cite{CL+2019} 
& A 
& APP & any & TFunc
& LO & TF & Yes &
Yes &  
Re & Si & 
& - & - & - \\

\cite{NC+2022} 
& A 
& APP & any & TFunc
& LO/P & TF & Yes &
Yes &  
Re & Si & 
& - & - & - \\

\cite{LR+2022} 
& A 
& APP & any & TFunc
& LO  & KE & Yes & Yes 
& Re & Si & 
& - & - & - \\ 

\cite{MA+2020} 
& A 
& APP & any & PFunc/TFunc
& P/NO & PT & Yes & Yes  
& Re & Si & 
& - 
& - & - \\

\nuovo{\cite{RG+2023}}
& \nuovo{A} 
& APP & any & SA
& P & - & No & No  
& Re & Si & 
& - 
& - & - \\ 

\cite{GS+2020} 
& A 
& APP & any & PFunc/TFunc
& LO & PT & No & No  
& Re & Si & HG
& - 
& - & - \\

\cite{TL2021} 
& A 
& APP & any & PFunc/TFunc
& LO/P/NO & KE/TF & No & No  
& Re & Si & 
& - 
& - & - \\

\cite{SG+2019} 
& A 
& APP & any & TFunc
& DV & PT & No &  No
& Re & Si & 
& - & - & - \\

\cite{RS+2021} 
& A 
& APP & any & SA
& P & DK & No &
No &  
Re & Si & 
& - & - & - \\

\nuovo{\cite{GR+2023_2}}
& \nuovo{A} 
& APP & any & SA
& P & -- & No &
No &  
Re & Si & 
& - & - & - \\

\cite{ZI+2022_1} 
& A 
& APP & any & PFunc/TFunc
& P/LO & - & No & No  
& Re & Si & 
& - 
& - & - \\

\cite{LR+2021} 
& A 
& DEV & FPGA & SEU/SET
& REG & TF & No & No 
& Re & Ra & HG
& - & - & - \\

\cite{SP+2021} 
& A 
& RTL & FPGA & SEU
& REG & FI & No & No 
 &  
VF & Em & 
& - & - & - \\

\cite{WW+2021}
& A 
& DEV/RTL & FPGA & SEU
& REG & - & No & No 
 &  
Re & Em & 
& - & - & - \\

\cite{XZ+2021}
& A 
& RTL & FPGA & SA
& REG & - & Yes & Yes 
 &  
Re & Em & 
& - & - & - \\

\nuovo{\cite{GG+2023}}
& \nuovo{A} 
& RTL & FPGA & SEU
& REG & - & No & No 
& Re & Em & HG
& - & - & - \\

\nuovo{\cite{TI+2023}} 
& \nuovo{A} 
& RTL & FPGA & SEU
& REG & - & No & No 
& Re & Em/Ra & HG
& - & - & - \\

\cite{RM+2022} 
& A 
& DEV & TPU & SEU
& REG & TF & No & No 
& Re & Ra & 
& - & - & - \\

\cite{CC+2021_1} 
& A 
& RTL & TPU & SA
& REG & PT & No & No  
& Re & Si & 
& - 
& - & - \\

\nuovo{\cite{PR+2023}}
& \nuovo{A} 
& ISA & TPU & SA
& P/DV & ND & No & No  
& Re & Si & 
& - 
& - & - \\

\cite{HBL2020} 
& A 
& RTL/APP & TPU & SEU
& REG & TF & Yes & Yes
& Re & Si & HG
& - 
& - & - \\

\cite{LF+2021} 
& A 
& ISA/APP & CPU & SEU & REG  
& - & No &  No
& VF & Si & 
& - & - & - \\

\cite{MR+2021_1} 
& A 
& DEV/APP & CPU & SEU/SA
& REG & ND & No & No  
& Re & Ra/Si & 
& - 
& - & - \\

\cite{CG+2022} 
& A 
& ISA/APP & GPU & SA
& REG  & - & No & No 
& Re & Si & 
& - & - & - \\ 

\cite{CC+2021} 
& A 
& RTL/APP & FPGA & SEU
& REG & -  & No & No  
& Re & Em & 
& - 
& - & - \\

\cite{BC+2022_2} 
& A 
& ISA/APP & GPU & SEU
& REG & TF & Yes & Yes   
& VF/Re & Em/Si & 
& - 
& - & - \\

\cite{BC+2023}
& \nuovo{A} 
& ISA/APP & GPU & SEU
& REG & TF & No & No  
& Re/VF & Em/Si & HG
& - 
& - & - \\ 

\cite{HK+2023} 
& \nuovo{A} 
& RTL/APP & TPU & SEU
& REG & PT & No & No  
& VF & Si & HG
& - 
& - & - \\ 

\nuovo{\cite{TW+2023}}  
& \nuovo{A} 
& RTL/APP & TPU & SEU/SA  
& REG & KE & Yes & Yes  
& VF/Re & Em & HG 
& - & - & - \\ 

\cite{DAS2022} 
& A 
& RTL/APP & FPGA & SEU
& REG & PT & No & No  
& Re & Em & 
& - & - & - \\

\cite{ACP2022} 
& A 
& ISA/APP & CPU & SEU & REG  
& any & Yes  & Yes  
& Re/VF & Si & HG
& - & - & - \\ 

\cite{TR+2023} 
& \nuovo{A} 
& RTL/APP & FPGA & SEU
& REG & KE & No & No  
& Re & Si & 
& - & - & - \\ 

\cite{LH+2019} 
& A 
& DEV/RTL/APP & GPU & SEU
& REG & TR & No & No 
 &
Re/Sa & Ra/Si & 
& - & - & - \\

\cite{NA+2019} 
& A 
& APP & any & TFunc
& P & CA & Yes &  Yes
& Re & Si & HG
& - & - & - \\

\cite{IWA2020} 
& A 
& ISA/APP & GPU & SEU  
& REG & DK & No & Yes  
& VF/Re & Em & 
& - & - & - \\ 

\cite{MR+2021} 
& A 
& APP & any & TFunc
& P & - & No & No  
& Re & Si & HG
& - 
& - & - \\ 

\cite{KB+2021}
& \nuovo{A} 
& RTL/APP & TPU & SA
& REG & KE & No & No  
& Re/VF/Sa & Si & 
& - 
& - & - \\ 



\hline\hline
\cite{SHT2021} 
& H 
& APP & any & SEU/SA
& P/DV & PT & Yes & Yes 
& - & - & -
& FT & Sel & TMR/ECC \\

\cite{GG2021} 
& H 
& ALG & GPU & SEU
& REG/LO & TF & No &
No &  
- & - & - &
FT & Ex & TMR \\

\cite{RG+2022} 
& H 
& APP & any & TFunc 
& NO & PT & No & No  
& - & - & -
& FT & Sel & TMR \\

\cite{GZ+2022} 
& H 
& APP & FPGA & SEU
& REG & - & No & No  
& - & - & -
& FT & Ax & NMR \\

\cite{DW+2021} 
& H 
& RTL & TPU & SEU
& REG & - & No & No 
& - & - & -
& FT 
& Sel & D+R \\

\cite{ZD+2021} 
& H 
& ALG & any & SEU 
& LO & CF & No & Yes  
& - & - & -
& FD/FT & Spec & ABED/ABFT \\

\cite{HS+2022}  & H 
& ALG & GPU & SEU
& REG/LO/P & cu & No & No  
& - & - & -
& FD 
& Spec & \ac{ABED}/DWC \\

\cite{KR2021} 
& H 
& ALG & GPU & SEU
& DV & - & No & No  
& - & - & -
& FD 
& Spec & \ac{ABED} \\ 


\cite{OO2020_1} 
& H 
& ALG & any  & SEU 
& P/LO & KE/TF & No & No  
& - & - & - 
& FD & Spec & \ac{ABED} \\ 

\cite{CLP2021} 
& H 
& APP & any & SEU
& DV & TF & Yes &
Yes & 
- & - & - &
FT & Full & DL \\

\cite{HHS2020} 
& H 
& APP  & any & Pfunc
& P & PT & No & Yes 
& - & - & -
& FT 
& Full & DL \\

\cite{GS+2022} 
& H 
& APP & any & SEU
& P & PT & No &
No & 
- & - & - &
FT & Full & DL \\

\cite{OO2021} 
& H 
& APP  & any & Pfunc
& P/LO & PT & No & No 
& - & - & -
& FT 
& Full & DL \\

\cite{AM+2022} 
& H 
& APP  & any & Pfunc
& P/LO & PT & No & No 
& - & - & -
& FT 
& Full & ABEF/DL \\

\cite{GS+2022_1} 
& H 
& APP  & any & SA
& P & - & No & No 
& - & - & -
& FT 
& Full & DL \\

\cite{GN+2019} 
& H 
& APP & any & SEU  
& P & PT & No & No  
& - & - & - 
& FT & Spec & \ac{ECC}/\ac{DL} \\

\cite{LY2022_1} 
& H 
& APP & any & SA  
& P & KE & No & No  
& - & - & - 
& FT & Spec & \ac{ECC}/\ac{DL} \\

\cite{ZG+2020}
& \nuovo{H} 
& APP & GPU & SA
& P & PT & No & No  
& - & - & -
& FT & Full & DL\\

\cite{CS+2022}
& H 
& APP & \ac{GPU} & SEU
& REG/LO & - & No & No  
& - & - & -
& FT & Full & DL\\

\cite{BS+2022}
& H 
& APP & any & SEU
& P/NO & - & No & No  
& - & - & -
& FT & Full & DL\\

\nuovo{\cite{HY+2023}}
& \nuovo{H} 
& ALG & TPU & SA
& P/LO & - & No & No  
& - & - & -
& FT & Spec & ABFT\\

\hline
\hline

\cite{PTY2020} 
& B 
& APP & any & TFunc
& P & KE/TF & No &
No &  
VF/Re & Si & 
& FD/FT & Sel & ECC+DWC \\

\cite{FdS+2019} 
& B 
& DEV/ISA/ALG & GPU & SEU 
& REG & DK & No & No  
& Re & Ra/Em & 
& FT & Spec & \ac{ECC}/\ac{ABFT}\\ 

\nuovo{\cite{HH+2023}} 
& B 
& RTL/APP & TPU & SEU
& REG & TF & Yes &
Yes &  
Re & Si & HG 
& FT & Spec & D+R \\

\cite{IWB+2019} 
& B 
& ISA/APP & GPU & SEU  
& REG & DK & No & Yes  
& VF/Re & Em & 
& FT & Sel & TMR \\ 

\cite{HL2023} 
& \nuovo{B} 
& RTL/APP & TPU & SA  
& REG & - & No & No  
& VF & Si & HG 
& FD/FT & Full & DL \\

\cite{MH+2021} & B 
& APP & any & \ac{SEU} & NO 
& PT & No & No 
& VF & Si & 
& FT
& Sel & D+R/\ac{ABED} \\ 

\cite{AG+2021} 
& B 
& ISA/ALG & CPU & SEU  
& REG & CN & No & No  
& Rel & Si & 
& FT & Sel & TMR/DL \\

\cite{BC+2022_1} 
& B 
& ISA/APP & GPU & SEU
& LO & TF & No &
Yes & 
VF/Re & Em/Si & 
& FD & Sel & \ac{DWC} \\

\cite{KR+2020} 
& B 
& APP & FPGA & SEU 
& P & FI & No & No  
& VF & Em & 
& FT & Sel & TMR \\

\cite{BG+2022} 
& B 
& APP & FPGA & SA  
& P & FI & No & No  
& VF & Em & 
& FT & Sel & TMR \\

\cite{TKS2023} 
& \nuovo{B} 
& APP & any & SEU
& P & - & No & No &  
VF & Em & HG 
& FT & Sel & \ac{ECC} \\

\cite{YS+2020}
& \nuovo{B} 
& APP & any & SEU
& P & - & No & No &  
VF & An & HG 
& FT & Sel & \ac{ECC}/\ac{TMR} \\

\cite{BEA2021} 
& B 
& APP & any & SEU  
& P & ND & No & Yes  
& Re & Si & 
& FD/FT & Spec & \ac{ECC}/\ac{DL} \\ 

\cite{SH2023} 
& B 
& RTL/APP & TPU & SA
& REG & TF & No &
No & 
VF/Re & Si & 
& FT & Full & \ac{DL} \\

\cite{LL+2021} 
& B 
& ISA/APP & CPU & SEU  
& REG & CM & No & No  
& Re/VF & Em & 
& FT & Sel & D+R/\ac{DL}/CHK \\ 

\cite{ZH+2019} 
& B 
& APP & any & SEU
& P/DV & CF & No &
No &  
Re & Si & 
& FD & Spec & \ac{ABED} \\


\hline

\end{longtable}
\end{landscape}

\twocolumn

\begin{table*}[h!]
\caption{Open-source software made available from the works presented in Table~\ref{tab:all} (Tool support)}\label{tab:opensw}
\begin{tabular}{lll}
\textbf{Ref.}    & \textbf{Name}   & \texttt{url} 
\\\hline\hline
\cite{RG+2018} & Ares & alugupta.github.io/ares \\
\cite{CL+2019} & BinFI & github.com/DependableSystemsLab/TensorFI-BinaryFI \\ 
\cite{NC+2022} & TensorFI2 & github.com/DependableSystemsLab/TensorFI2 \\
\cite{LR+2022} & TensorFI+ & github.com/sabuj7177/characterizing\_DNN\_failures \\
\cite{MA+2020} & PyTorchFI & github.com/PyTorchfi/PyTorchfi \\
\cite{XZ+2021} & & github.com/ICT-CHASE/fault-analysis-of-FPGA-based-NN-accelerator \\
\cite{HBL2020} & FIdelity & github.com/silvaurus/FIdelityFramework \\
\cite{HH+2023} & \nuovo{FIdelityTraining} & https://github.com/YLab-UChicago/ISCA\_AE \\
\cite{BC+2022_2} & CLASSES & github.com/D4De/classes\\
\nuovo{\cite{TW+2023}} & \nuovo{saca-FI} & github.com/One-B-Tree/Saca-FI-A-microarchitecture-level-fault-injection-framework-for-CNN-accelerator \\
\cite{ACP2022} & LLTFI & github.com/DependableSystemsLab/LLTFI \\
\cite{NA+2019} & & github.com/cypox/CNN-Fault-Injector \\
\cite{SHT2021} & & github.com/Msabih/FaultTolerantDnnXai \\
\cite{CLP2021} & Ranger & github.com/DependableSystemsLab/Ranger \\
\hline
\end{tabular}
\end{table*}

\section{Insights, challenges and opportunities}\label{sec:insights}
The high number of pertinent contributions in the last four years (i.e., \numpre{} authored by more than 400 scientists) shows a dynamic context, that in this decade has been fostering  interesting and relevant outcomes, characterized by some common aspects, that we summarize in the following, together with open challenges and opportunities (beyond the ones highlighted by~\cite{WS+2022}).

\begin{figure*}[htb]
    \centering
    \includegraphics[width=0.9\textwidth]{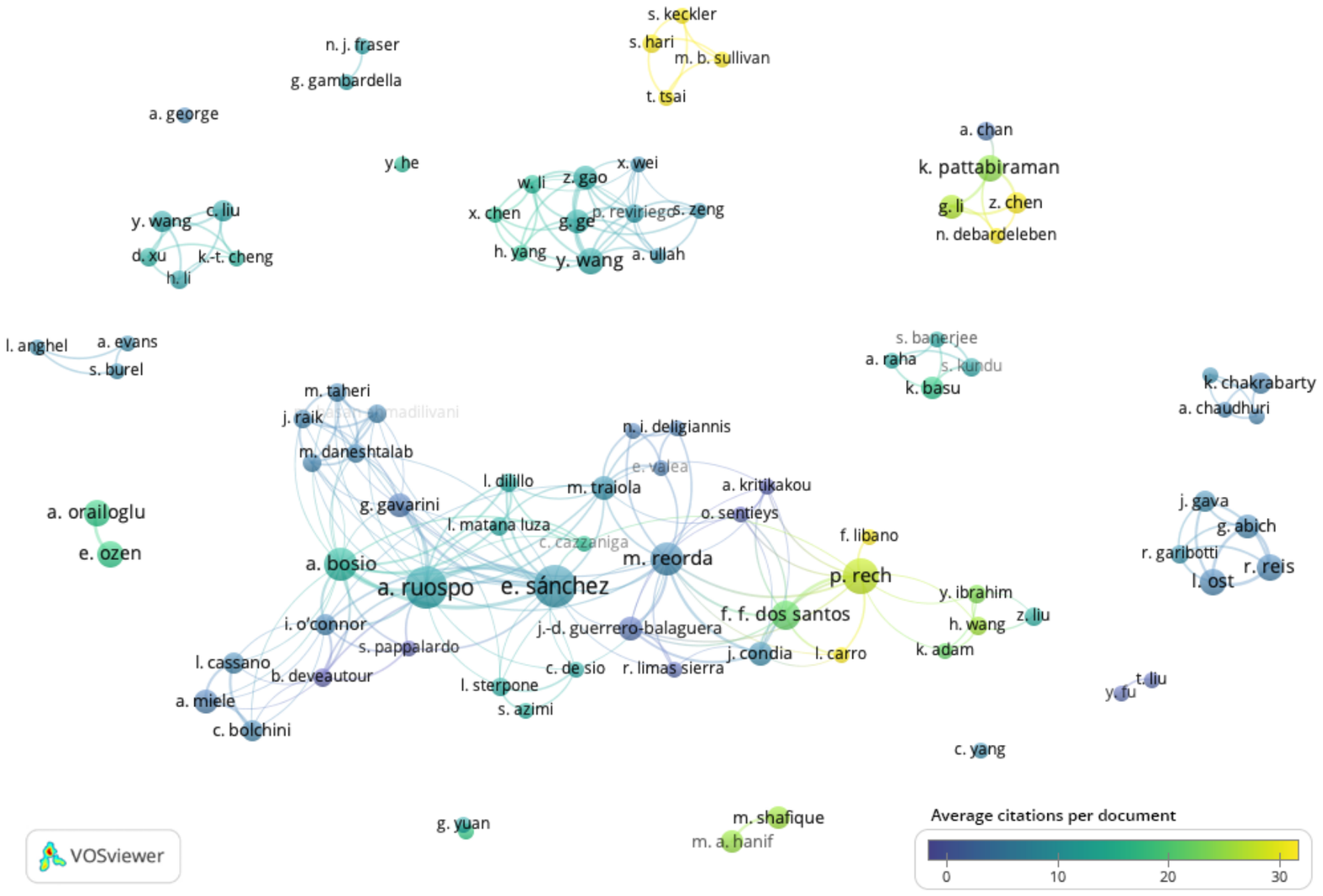}
    \caption{Co-authorship analysis with ``authors'' as the unit of analysis. In this analysis, the minimum number of documents for each author is 3, and the number of selected authors is 93, grouped in 14 clusters, accordingly. Node size depends on the number of documents and the connecting lines between them indicate the collaboration between authors. The color spectrum represents the average number of citations.}
    \label{fig:coauthors}
\end{figure*}

\begin{figure*}
\centering
\begin{tabular}{c}
   \includegraphics[width=0.8\textwidth]{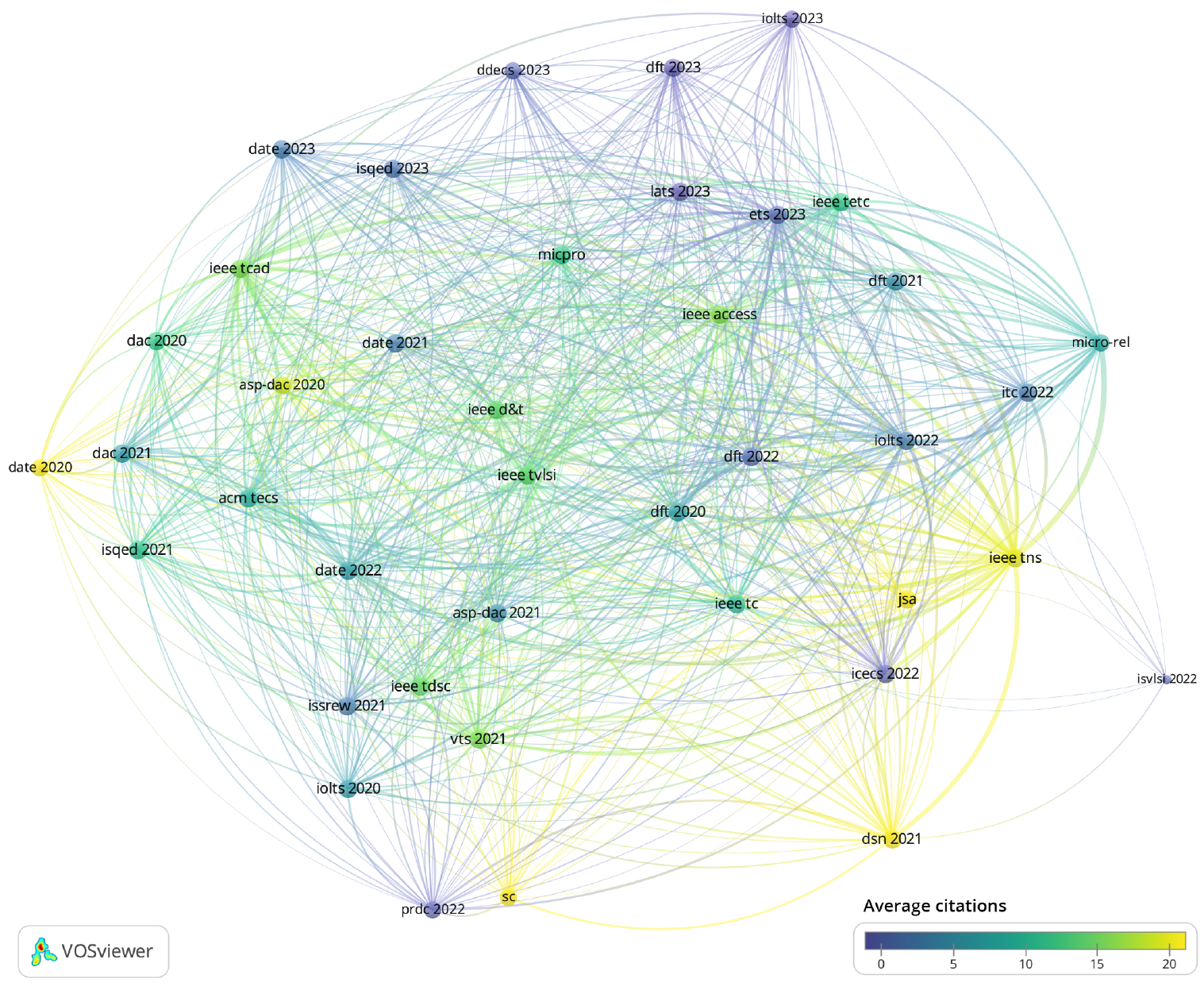} \\
   (a) \\
   \includegraphics[width=0.8\textwidth]{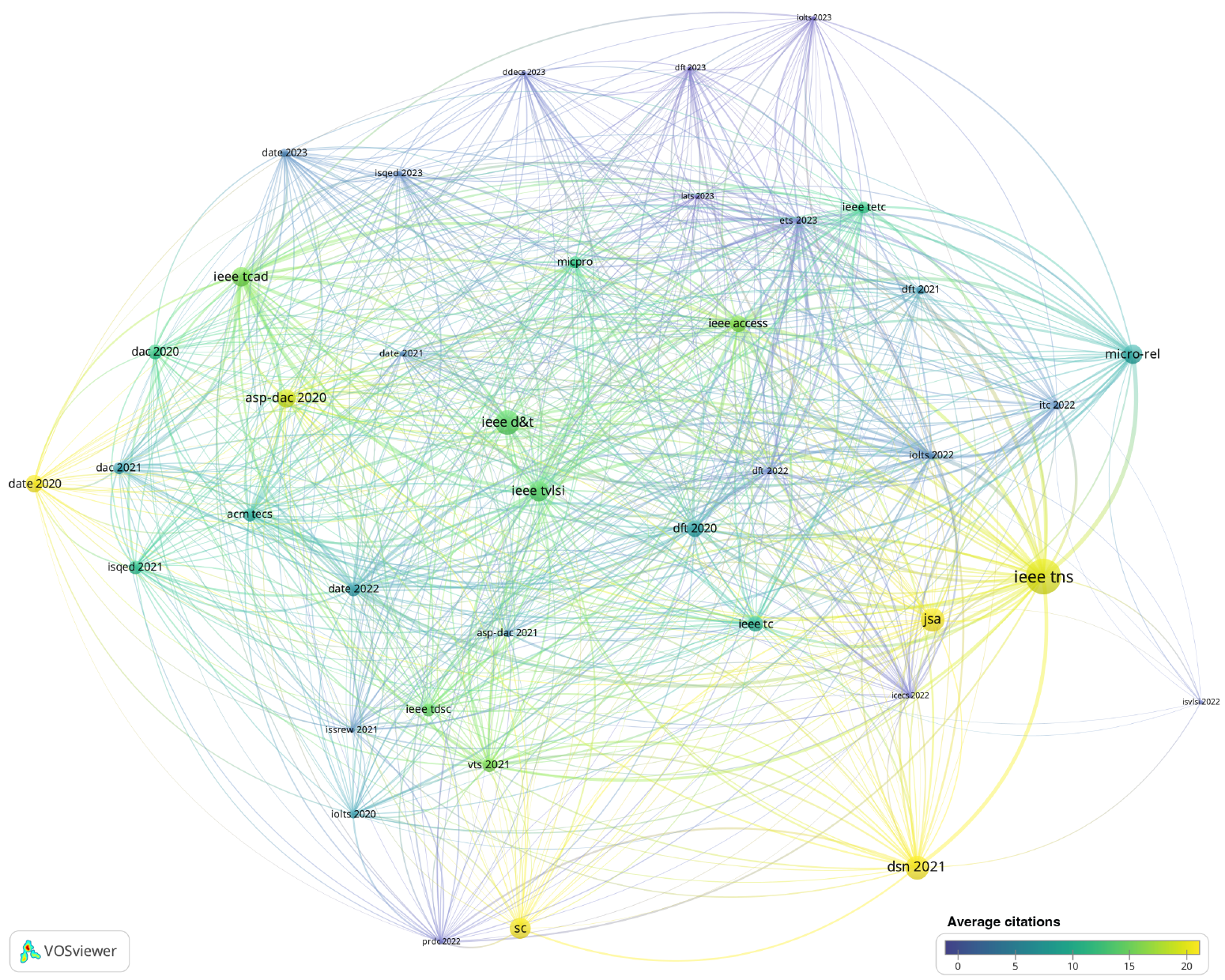} \\
   (b) \\
\end{tabular}
  \caption{Eligible studies: analysis of the publication venues with respect to (a) the number of papers at such a venue and the (b) cross-reference among them. A link between two items means that one of them cites the other and the color spectrum represents the average number of citations.}
        \label{fig:venue}
\end{figure*}

\begin{description}
\item[Trend] The number of contributions in the years has been increasing (as Figure~\ref{fig:timeline} shows) if we consider that the spectrum of analysis and design targets has grown and the works reported in the chart cover only a limited research area (the one included in this survey) with respect to the total. 

\item[\ac{DL} design impact on resilience] Numerous are the studies that explore how different \ac{DL} design choices -- from data type, to data quantization, from pruning to compression -- affect the resulting network resilience to faults corrupting both stored data (e.g., weights, neuron output) and manipulation (e.g., convolution output). Such impact, though, is heavily and strictly related to the specific adopted \ac{DL} solution, and although some general considerations are drawn, there is no ``one ground truth that applies to every case'' so that, in our opinion, every time a \ac{DL} application has to be deployed in a safety/mission-critical application domain, analysis and hardening solutions need to be specifically tailored. To this end, approaches providing usable tools and methods to analyze and harden a \ac{DL} application seem to be of great interest.
\item[Metrics] For both the analysis and hardening strategies, most contributions can be partitioned into two classes, those evaluating resilience with respect to conventional reliability metrics, such as \acf{MTTF}, \acf{FIT}, \acf{AVF}, \acf{PVF}, \acf{KVF} or the \ac{SDC} rate (e.g., \cite{FdS+2019}) and those who adopt an \textit{application-aware} metric, more closely related to the specific and special context, such as usable/not usable (e.g., \cite{BB+2022, AG+2021}). Both classical and innovative figures of merit are adopted or defined, leading to numerous alternative visions.  Some of the best contributions report comparative results that allow the reader to identify benefits and potentials of the new discussed solutions, but the rich set of different quantitative metrics makes the task not an easy one. \textbf{Challenge:} Although the choice of the adopted metric depends on the application context, future efforts could go in the direction of reporting always the results also with respect to a commonly adopted metric, to enable fair comparisons.
\item[Cross-layer strategies] The complexity of the hardware platforms able to efficiently execute heavy \ac{ML}/\ac{DL} applications and that of the applications themselves initially led to contributions that worked either at the architecture level (working on faults) or at the application level (working on errors). However, the gap between these levels and the necessity to maintain a correspondence between faults and errors to provide a reliable susceptibility/resilience evaluation are spurring cross-layer approaches that explore and support such a fault-error relation.
\item[Fault injection tools and their availability] Considering the application context and the involved elements, fault injection is a critical task with respect to 
\begin{inparaenum}[i)] 
\item the experiment time, 
\item the controllability/observability aspects, and 
\item the adherence of the injected errors to the underlying realistic faults.
\end{inparaenum} 
 Specifically targeting the domain of interest, several fault injection tools have been recently proposed, working from the architectural level (\cite{CL+2019,HBL2020,TH+2021,ACP2022}) to the application one (\cite{RG+2018,NC+2022,MA+2020}), or cross-layer (\cite{DAS2022,BC+2022_2}). Although several of them are available (see Table~\ref{tab:opensw} for the available open-source software packages), when developing hardening techniques and strategies, proprietary fault injection solutions are devised, sometimes to drive a selective hardening policy based on the analysis outcomes. 
\textbf{Challenge:} an ecosystem of available tools working at different abstraction levels, on different platforms could indeed allow for a systemic effort to tackle \ac{DL} resilience for present and future challenges.

\item[Reproducible research] One of the critical activities when developing new methods is the evaluation of their performance with respect to existing ones, to motivate the introduction of yet another approach. Often, the comparison is carried out against the vanilla solution, the baseline implementation without any sort of hardening. Indeed, only a few contributions (besides the ones proposing a new tool) share and make public their software/data. \textbf{Challenge:} incentive reproducible research to foster stronger contributions, as well as the possibility to move towards an integrated ecosystem of solutions for the different hardware/software/application variants. As an example, an available benchmark suite that offers for the various hardware/software/application contexts a reference to \begin{inparaenum}\item compare solutions, and \item support the integration of complementary approaches \end{inparaenum}, could be a valuable asset for the community.
\item[Community] There are a number of very active research groups on the topic, that are steadily contributing to the discussion. To visually get an overview of such a community, the awareness and relationships among the research groups, as well as the typical venues where the topic is presented and discussed, we exploited VOSviewer (\cite{vEW2010}).
On the \numread{} papers considered eligible we explored co-authorship and (see Figure~\ref{fig:coauthors}). The analysis identifies 68 authors having authored at least 3 papers on the topic, belonging to 14 clusters (research groups). Links between nodes represent a co-authorship. On the same dataset we explore the publication venues; the graphs in  Figure~\ref{fig:venue}(a) and~\ref{fig:venue}(b) show the venues where the included contributions have been published, highlighting the number of documents and the cross-references among venues, respectively.
Finally, we analyse the set of included papers (\numincluded{} in total) exploring the number of citations to possibly get insights on other scientists' awareness, reported in  Figure~\ref{fig:citations}.
\item[Synergy opportunity] This work, as well as past literature review analyses, shows that \ac{ML} resilience, and \ac{DL} in the specific, against faults affecting the underlying hardware is a research area exhibiting many challenges and facets, setting an opportunity for creating a synergy in the research community towards the development of an ecosystem of methods and tools that can tackle the different facets of \ac{DL} resilience against hardware faults. 
\end{description}

\begin{figure*}[htb]
    \centering
    \includegraphics[width=\textwidth]{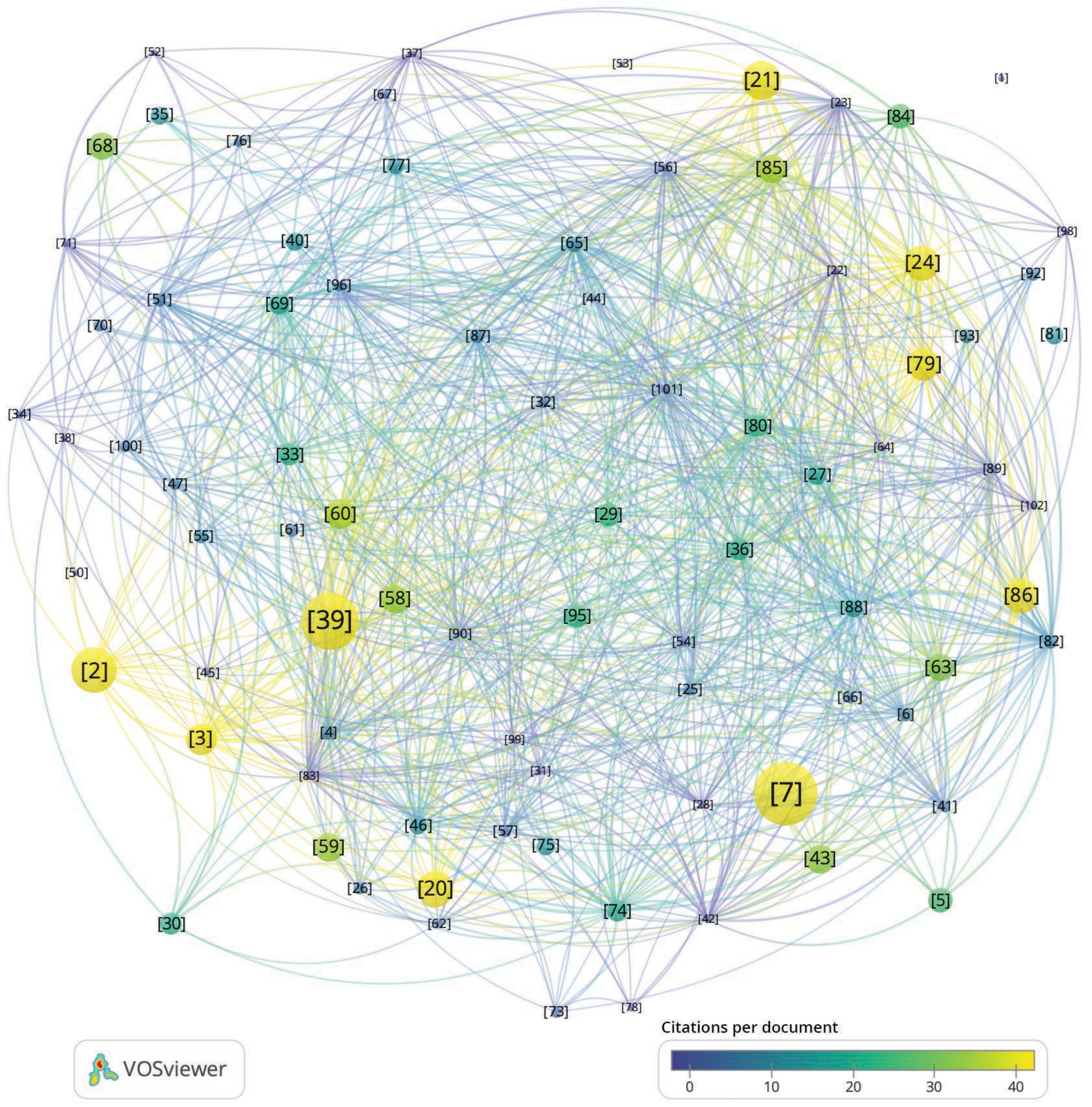}
    \caption{Included studies: citations counts indicating the most cited literature and the cross-reference among them. Node size depends on the number of citations and the connecting lines between them indicate the reference in the bibliography.}
    \label{fig:citations}
\end{figure*}

\section{Concluding remarks}\label{sec:end}
This paper collects and reviews the most recent literature (since 2019) on the analysis and design of resilient \ac{DL} algorithms and applications against faults in the underlying hardware. The analysis includes \numincluded{} studies focused on methods and tools dealing with the occurrence of transient and permanent faults possibly causing the \ac{DL} application to misbehave. Through a detailed search and selection process we reviewed \numincluded{} contributions, and analyzed them with respect to a classification framework supporting the reader in the identification of the most promising works based on the area of interest (e.g., with respect to the adopted fault model, error model or \ac{DL} framework). The aim is twofold; i) mapping the active research landscape on the matter, and ii) classifying the contributions based on various parameters deemed of interest to support the interested reader in finding the relevant information they might be looking for (e.g., similar studies, solutions that might be applied, etc.). 
The study emphasizes the breadth of the research and actually defines some boundaries to limit the included contributions, focusing on \ac{DL} applications and the most commonly adopted fault models, leaving other facets (e.g., spiking neural networks, \nuovo{vision transformers}, manufacturing and process-variation faults) to future studies.
Some insights and overall considerations are also drawn; the vibrant research on this topic and the broad spectrum of challenges calls, in our opinion, towards the development of an ecosystem of solutions that offer a support in the implementation of resilient \ac{DL} applications. 

\section*{CRediT authorship contribution statement
}
\textbf{Cristiana Bolchini:} Conceptualization, Methodology, Investigation, Data Curation, Writing - Original Draft, Writing - Review \& Editing, Visualization. \textbf{Luca Cassano:} Conceptualization, Methodology, Investigation, Writing - Original Draft, Writing - Review \& Editing. \textbf{Antonio Miele:} Conceptualization, Methodology, Investigation, Writing - Original Draft, Writing - Review \& Editing.

\section*{Declaration of competing interest}
The authors declare that they have no known competing financial interests or personal relationships that could have appeared to influence the work reported in this paper.

\section*{Data availability}
No specific data was used for the research, except for the lists of papers matching the search queries.


\end{document}